\documentclass[11pt, letterpaper, shortlabels]{berkeley}

\usepackage{algorithm2e}
\usepackage{algorithm}
\usepackage{algorithmicx}
\usepackage{algpseudocode}

\usepackage{dsfont}
\usepackage{nicefrac}
\usepackage{subcaption}
\usepackage{amsmath, amssymb}
\usepackage{enumitem}
\usepackage{textcomp}
\usepackage{lscape}

\usepackage{url}
\usepackage{enumitem}
\usepackage{booktabs} 
\usepackage{array}     
\usepackage{multirow}
\usepackage[utf8]{inputenc}
\usepackage{tabularx, makecell}
\usepackage[table]{xcolor}   
\usepackage{caption}
\usepackage{graphicx}
\usepackage{wrapfig}
\usepackage{listings}
\usepackage{soul}
\usepackage[most]{tcolorbox}
\usepackage{fontawesome5}

\usepackage{pifont}    
\definecolor{rowgray}{gray}{0.95}
\usepackage[T1]{fontenc}

\usepackage{xcolor}             
\usepackage{threeparttable}
\usepackage{makecell}

\definecolor{rowblue}{RGB}{230,245,255}
\definecolor{rowgreen}{RGB}{235,250,235}
\definecolor{roworange}{RGB}{255,245,230}
\definecolor{rowyellow}{RGB}{255,255,230}
\definecolor{rowpurple}{RGB}{245,235,255}
\definecolor{lightblue}{RGB}{220,235,247}

\definecolor{tableHeaderBg}{RGB}{230,230,230} 
\definecolor{tableRow}{RGB}{245,245,245}

\newtcblisting[auto counter, number within=section]{PromptBox}[2][]{%
  enhanced,
  breakable,
  sharp corners=all,
  colback=blue!3,                             
  colframe=blue!20,                           
  boxrule=0.8pt,
  arc=4mm,                                    
  borderline west={3mm}{0pt}{blue!15},        
  colbacktitle=blue!8,                        
  coltitle=black,
  fonttitle=\bfseries,
  attach boxed title to top center={yshift=-3mm}, 
  boxed title style={
    sharp corners, 
    boxrule=0.6pt,
    colback=blue!12,
    colframe=blue!25
  },
  title={#2}, 
  before skip=15pt, 
  after skip=15pt,
  top=5mm, 
  bottom=5mm, 
  left=6mm, 
  right=6mm,
  drop shadow={black!5!white},                
  listing only,
  listing engine=listings,
  listing options={
    basicstyle=\ttfamily\footnotesize,
    breaklines=true,
    columns=fullflexible,
    keepspaces=true,
    showstringspaces=false,
    tabsize=2,
    breakindent=0pt,
    breakatwhitespace=true,
    backgroundcolor=\color{blue!2}            
  },
  #1
}

\newcolumntype{P}[1]{>{\centering\arraybackslash}p{#1}}

\DeclareUnicodeCharacter{22A5}{\ensuremath{\bot}} 
\DeclareUnicodeCharacter{00AC}{\ensuremath{\neg}} 
\DeclareUnicodeCharacter{2227}{\ensuremath{\land}} 
\DeclareUnicodeCharacter{2228}{\ensuremath{\lor}} 
\DeclareUnicodeCharacter{2192}{\ensuremath{\rightarrow}} 
\DeclareUnicodeCharacter{2203}{\ensuremath{\exists}} 

\newtcolorbox{bluebox}[1][]{
  enhanced,
  colframe=blue!40!gray,
  colback=white,
  coltitle=white,
  colbacktitle=blue!40!gray,
  width=\linewidth,
  arc=2mm,
  auto outer arc,
  boxrule=0.5pt,
  left=10pt,
  right=10pt,
  drop shadow={black!50!white},
  top=10pt,
  bottom=10pt,
  title={#1}, 
  fonttitle=\bfseries,
  title code={\node[rounded corners, fill=blue!75!black, draw=none, text=white] at (frame.title) {\textbf{#1}};}, 
  attach boxed title to top center={yshift=-2mm},
  boxed title style={sharp corners, size=small}
}

\title{TellOnce: Correcting Coding Agents Once, Improving Their Future Actions}

\newcommand{\headertitle}{TellOnce: Correcting Coding Agents Once, Improving Their Future Actions}

\usepackage[all]{hypcap}

\usepackage[authoryear, round]{natbib}

\usepackage{hyperref}[citecolor=magenta,linkcolor=magenta]

\usepackage{multirow, makecell, caption}\usepackage{microtype}
\usepackage{graphicx}
\usepackage{booktabs} 

\usepackage{amsmath}
\usepackage{amssymb}
\usepackage{mathtools}
\usepackage{amsthm}
\usepackage{mathrsfs}
\usepackage{nicefrac}
\usepackage{dsfont}
\usepackage{enumitem}
\usepackage{arydshln}
\usepackage{xspace}
\usepackage[capitalize,noabbrev]{cleveref}
\usepackage{subcaption}
\usepackage{lipsum}
\usepackage{listings}
\usepackage{tcolorbox}
\usepackage{fontawesome5} 
\usepackage{amsmath}
\usepackage{amssymb}
\usepackage{mathtools}
\usepackage{amsthm}
\usepackage{bbm}
\usepackage{soul}
\usepackage{algpseudocode}
\usepackage{setspace}

\makeatletter
\def\mathcolor#1#{\@mathcolor{#1}}
\def\@mathcolor#1#2#3{%
  \protect\leavevmode
  \begingroup
    \color#1{#2}#3%
  \endgroup
}
\makeatother

\definecolor{NDblue}{RGB}{12, 35, 64} 
\definecolor{NDgold}{RGB}{174, 145, 66} 

\hypersetup{
    colorlinks = true,    
    linkcolor = NDblue,    
    citecolor = NDgold,   
    urlcolor = NDblue,     
    filecolor = NDblue     
}

\Crefformat{equation}{#2Eq.\;(#1)#3}

\Crefformat{figure}{#2Figure #1#3}
\Crefformat{assumption}{#2Assumption #1#3}
\Crefname{assumption}{Assumption}{Assumptions}

\usepackage{crossreftools}
\definecolor{c1}{HTML}{D8DFE5}
\definecolor{c2}{HTML}{A1B4A5}
\definecolor{c3}{HTML}{C5D3E3}
\definecolor{c4}{HTML}{849E8A}
\definecolor{c5}{HTML}{A5C4BD}
\definecolor{c6}{HTML}{EFF4F7}
\definecolor{c7}{HTML}{778FD2}
\definecolor{c8}{HTML}{f6f4f0}
\definecolor{ceruleanblue}{rgb}{0.16, 0.32, 0.75}

\author{%
  Yujun Zhou$^1$,
  Kehan Guo$^1$,
  Haomin Zhuang$^1$,
  Xiangqi Wang$^1$,
  Yue Huang$^1$,
  Zhenwen Liang$^3$,
  Pin-Yu Chen$^2$,
  Tian Gao$^2$,
  Nuno Moniz$^1$,
  Nitesh V. Chawla$^1$,
  Xiangliang Zhang$^1$ \\
  $^1$University of Notre Dame $^2$IBM Research $^3$Tencent AI Lab\\
  \texttt{\{yzhou25,xzhang33\}@nd.edu}
  }

\correspondingauthor{xzhang33@nd.edu (Xiangliang Zhang)}

\newcommand{\sysname}{\textsc{TellOnce}\xspace}

\definecolor{TellOnceAccent}{HTML}{176B6A}
\definecolor{TellOnceBand}{HTML}{EAF3F7}
\definecolor{TellOnceTint}{HTML}{E1F2EB}
\definecolor{ModelBand}{HTML}{EAF4F6}
\newcolumntype{Y}{>{\centering\arraybackslash}X}
\newcolumntype{T}{>{\columncolor{TellOnceTint}\centering\arraybackslash}X}

\begin{abstract}
\textbf{Abstract:}
Interactive LLM agents increasingly work with users across tasks, yet
remembering a previous correction does not ensure that the agent changes
its behavior in the future. In a diagnostic constructed from anonymized
real-world agent interactions, agents using Mem0
still violate 57.5\% of applicable preference checks.
We introduce \sysname{}, a drop-in personalization layer that turns
conversational corrections into persistent, scoped, and revisable
behavioral rules without updating the underlying agent model.
\sysname{} identifies which requirements should persist, determines
when they apply, reconciles subsequent feedback with existing rules,
and prepares reusable checks when reliable verification is possible.
We evaluate whether learned corrections improve future behavior by
freezing the learned state and testing on subsequent tasks without
additional user corrections. Across three models on ClawArena and
MemoryArena, \sysname{} achieves the lowest preference violation rate
in all six settings, reducing violations relative to PGMem by
19.3--42.3 percentage points on ClawArena and 3.0--22.9 points on
MemoryArena, while maintaining broadly comparable task performance.
On SWE-Together, \sysname{} further improves task quality
on new coding tasks by learning from feedback received
during earlier tasks. Experiment code is available at \url{https://github.com/YujunZhou/TellOnce_exp}, and the deployable skill at \url{https://github.com/YujunZhou/tellonce}.
\end{abstract}

\begin{document}
\maketitle

\section{Introduction}
\label{sec:intro}

Large language models are increasingly used as interactive agents that work
with users across tasks and sessions rather than as isolated question-answering
systems. They invoke tools, edit repositories, execute multi-step workflows,
and operate in environments whose state persists across extended
interactions~\citep{wang2025openhands,
he2026memoryarenabenchmarkingagentmemory,huang2026narra}.
As these systems become part of long-term human--AI collaboration,
personalization increasingly concerns not only what an agent knows about a
user, but also whether it adapts to how that user prefers to
work~\citep{salemi2024lamp,
zhao2025llmsrecognizepreferencesevaluating,
mehri2026multisessioncollablearninguserpreferences}.
If a user has already corrected the same behavior once, they should not
need to repeat that correction in every new task or session.

User corrections provide a particularly direct signal for such adaptation.
A user may say, for example, ``clean temporary files before finishing,'' or ``explain the important
tradeoffs when recommending an option.'' Unlike an ordinary user fact,
such feedback specifies how the agent should behave differently in the future.
However, not every correction should apply to every task: some express
general expectations, while others apply only to particular projects or
situations. The agent must therefore learn both what behavior should change
and when that change is relevant.

Prior work has learned user preferences from edits, interaction feedback,
conversation histories, and user summaries~\citep{gao2024aligning,
liang2026learningpersonalizedagentshuman,
nam2026plus,salemi2025naturalfeedback}.
Long-term memory systems have also made considerable progress in storing,
organizing, updating, and retrieving information across
sessions~\citep{mem0,xu2025amemagenticmemoryllm,
latimer2025hindsight2020buildingagent,cao2026remembermerefineme}.
These approaches make past user information increasingly accessible, but
remembering a correction does not guarantee that the agent will act on it.
An agent may retrieve a preference at task start but lose track of it
during a long execution, ultimately failing to follow it.
This raises a central question:
\textbf{How can an agent learn from a user's correction so that it changes
its behavior in relevant future tasks, without requiring repeated reminders?}

\begin{figure*}[t]
    \centering
    \includegraphics[width=.95\linewidth]{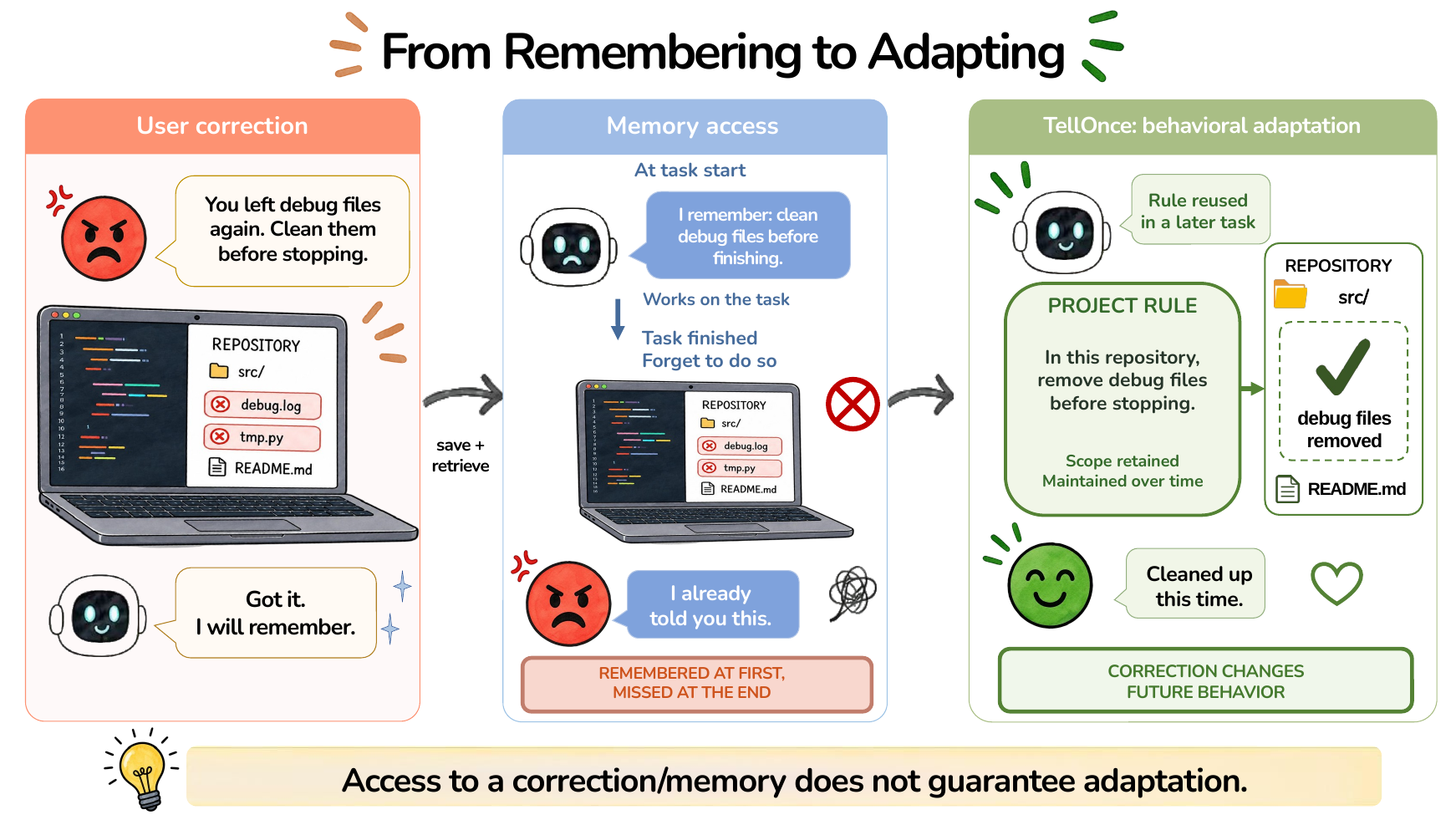}
    \caption{A past correction can be retrieved without becoming a stable
    change in future behavior. \sysname{} turns corrections into scoped
    behavioral rules for reuse across tasks.}
    \label{fig:memory-not-enough}
\end{figure*}

We first investigate this problem through a diagnostic constructed from
anonymized real-world coding-agent correction cases. With Mem0~\citep{mem0}, agents still violate
57.5\% of applicable preference checks. Providing natural-language rules directly
in context yields 55.0\% compliance, whereas representing the underlying
preferences as scoped rules with explicit checking criteria and using those
checks during execution yields 70.1\%. These findings reveal an
\emph{access--compliance gap}: making past feedback available is useful,
but does not ensure that the agent follows it. Reliable adaptation also
requires a way to carry the intended behavior into subsequent tasks and
assess whether it has actually been followed.

We introduce \sysname{}, a drop-in personalization layer that learns
reusable behavioral knowledge from conversational corrections without
updating the underlying agent model. Rather than simply storing each
correction as another memory item, \sysname{} determines whether it expresses
a requirement that should persist and identifies the behavior, scope,
conditions, and exceptions that need to be preserved. Because this knowledge
is reused across tasks, subsequent feedback may refine or replace what
was previously learned; \sysname{} reconciles such feedback with existing
rules instead of accumulating potentially conflicting memories.
In later tasks, relevant rules guide the agent's behavior. When reliable
checking is possible, \sysname{} also prepares reusable checks and uses
their results to identify violations and request revisions. Otherwise,
the learned rule remains available as behavioral guidance.

We evaluate \sysname{} end to end on ClawArena~\citep{
ji2026clawarenabenchmarkingaiagents} and the travel-planning tasks from
MemoryArena~\citep{he2026memoryarenabenchmarkingagentmemory}, using simulated
users who provide corrections during learning. After learning, the
long-term state is frozen and evaluated on test tasks without additional
user corrections. Across three task-solving models and both benchmarks,
\sysname{} achieves the lowest preference violation rate in all six settings.
Compared with PGMem, the strongest competing method on preference following,
\sysname{} reduces violations by 19.3--42.3 percentage points on ClawArena
and 3.0--22.9 points on MemoryArena, while maintaining broadly comparable
task-completion performance.

\textbf{Contributions.}
\textbf{(1)} We identify and quantify an \emph{access--compliance gap}:
agents may repeat previously corrected behavior even when relevant feedback
is available, motivating behavioral adaptation beyond memory retrieval.
\textbf{(2)} We introduce \sysname{}, a drop-in personalization layer that
turns conversational corrections into persistent, scoped, and revisable
behavioral knowledge, with reusable checking when reliable verification
is possible.
\textbf{(3)} We evaluate correction reuse across multiple agents and tasks,
demonstrating substantially fewer repeated preference violations with
generally comparable task performance.

\section{Related Work}
\label{sec:related}

\textbf{Personalized LLMs and Preference Learning.}
Prior work on personalization has explored several ways to infer what a user
prefers from past interaction. Some methods retrieve user profiles or histories
to condition generation~\citep{salemi2024lamp}, while others infer preferences
from user edits, summarize user information for personalized reward modeling,
or learn from natural-language feedback~\citep{gao2024aligning,
nam2026plus,salemi2025naturalfeedback}. More recent work studies how such
preferences can be updated across repeated or multi-session interactions
~\citep{liang2026learningpersonalizedagentshuman,
mehri2026multisessioncollablearninguserpreferences,zhang2026personaagent},
with benchmarks increasingly testing whether models can continue following
user preferences over long histories~\citep{
samuel2025personagym,ma2026syntheticinteractiondatascalable,
zhao2025llmsrecognizepreferencesevaluating,
guo2026realisticpersonalizationevaluatinglonghorizon}.
\sysname{} instead focuses on
what happens after a user explicitly corrects the agent: how that correction
becomes a durable behavioral rule with the appropriate scope, how later
feedback updates it, and how the rule is reused in future tasks.

\textbf{Long-Term Memory for LLM Agents.}
Long-term memory systems address a complementary problem: how information from
past interactions should be stored, organized, updated, and retrieved over
time. Mem0, A-MEM, and Hindsight develop general mechanisms for extracting and
retrieving interaction knowledge~\citep{mem0,xu2025amemagenticmemoryllm,
latimer2025hindsight2020buildingagent}; MemOS, RGMem, and preference-aware
memory update methods further study memory lifecycle and evolution
~\citep{li2025memosmemoryosai,
tian2026rgmemrenormalizationgroupinspiredmemory,
sun2025preferenceawarememoryupdatelongterm}. Other work improves how relevant
memories are associated or selected~\citep{zhang2026evokingusermemory,
xu2026multigranularitymemory}, while ReMe distills reusable procedural
experience~\citep{cao2026remembermerefineme} and MemoryArena evaluates whether
stored information supports later actions across sessions~\citep{
he2026memoryarenabenchmarkingagentmemory}.
\sysname{} does not aim to replace these general memory mechanisms. It focuses
on a particular kind of long-term information---user corrections about agent
behavior---and represents them as scoped, revisable behavioral rules rather
than only as memories to be retrieved later.

\section{Measuring the Access--Compliance Gap}
\label{sec:access-compliance}

Memory-based personalization aims to make relevant user
preferences available across tasks. Yet a correction can
be stored, retrieved, and presented to the agent while
its final response or workspace state still violates it. We refer to this disconnect between preference
\emph{access} and preference \emph{compliance} as the
\emph{access--compliance gap}.

\begin{figure*}[t]
    \centering
    \includegraphics[width=\linewidth]{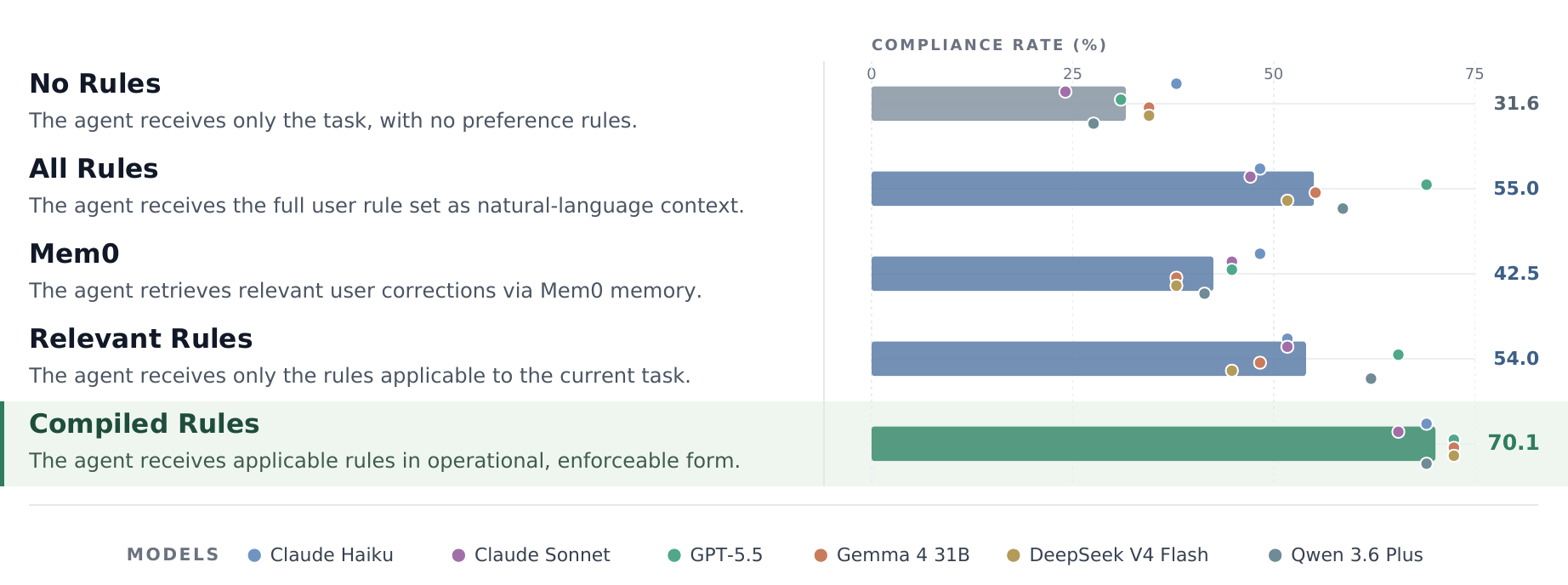}
    \caption{
    Diagnostic setup and results for the access--compliance gap.
    Left: five conditions that differ in whether the target preference is
    unavailable, provided as natural-language context, retrieved from memory,
    filtered to task-relevant rules, or represented as a scoped rule with
    explicit checking criteria. Right: compliance across six models and
    29 preference checks. Bars show mean compliance over the six models;
    points show individual models.
    }
    \label{fig:access-compliance}
\end{figure*}

\subsection{Diagnostic Setup}
\label{sec:access-compliance:setup}

We build a diagnostic benchmark from anonymized correction cases observed in
real-world coding-agent interactions. Analyzing 32 long-context coding-agent
transcripts---each containing roughly one million tokens of interaction
history---we extracted 142 correction-conflict records: instances in which a
user correction can be paired with an agent behavior that violated it. From
these records, we hand-curate 19 held-out evaluation tasks using three
filters---repetition, deduplication, and self-contained context---with full
criteria in Appendix~\ref{sec:appendix:access-stores}. The target preference
is removed from each task prompt, so the agent receives the coding request and
surrounding context but not the preference being evaluated.

The 19 held-out tasks contain 29 manually annotated preference checks spanning
cleanup requirements, workflow constraints, and stylistic conventions. The
checks were labeled by one author and independently audited by a second, with
disagreements resolved by discussion before finalization. A check is satisfied
only when the agent's final response or workspace state follows the annotated
preference. The preference stores used by the diagnostic---a 29-rule
natural-language store, a 47-entry operational rule library, and a Mem0
store---are constructed only from the remaining records, as detailed in
Appendix~\ref{sec:appendix:access-stores}.

The diagnostic compares five ways of making past preferences available in a
later task, illustrated in the left panel of
Figure~\ref{fig:access-compliance}. \textsc{No Rules} provides only the task
prompt. \textsc{All Rules} places the full natural-language rule set in
context. \textsc{Mem0} stores historical correction contexts and retrieves
relevant memories for the current task~\citep{mem0}. \textsc{Relevant Rules}
provides only the natural-language rules selected as relevant to the task.
Finally, \textsc{Compiled Rules} uses the same applicable preferences as
\textsc{Relevant Rules}, but represents them as scoped behavioral rules with
explicit applicability and checking criteria. The checker can then evaluate
the agent's resulting behavior against those criteria.

This comparison tests whether explicit applicability
criteria and execution-time checking improve preference
following beyond retrieval alone. In particular,
\textsc{Relevant Rules} and \textsc{Compiled Rules}
use the same task-relevant preferences, but the latter
adds structured applicability and checking criteria
and uses these criteria to evaluate the agent's behavior.
The end-to-end process for learning and reusing such
rules is evaluated in Section~\ref{sec:experiments}.

\subsection{Access Is Not Enough}
\label{sec:access-compliance:result}

Figure~\ref{fig:access-compliance} reports compliance rates across six
models.
Under \textsc{No Rules}, agents satisfy 31.6\% of the applicable preferences.
Providing the complete natural-language rule set with \textsc{All Rules}
raises compliance to 55.0\%, but still leaves nearly half of the applicable
preferences violated. \textsc{Mem0} achieves 42.5\%, while selecting only the
task-relevant natural-language rules yields 54.0\%. In contrast,
\textsc{Compiled Rules} reaches 70.1\%.

These results show that access alone is insufficient.
Selecting relevant rules offers little improvement over
providing the full rule set (54.0\% versus 55.0\%),
whereas scoped rules with execution-time checking
reach 70.1\%. This suggests that preference following
benefits from making retrieved requirements actionable.

Qualitatively, the remaining failures follow a common pattern. The agent may
acknowledge the correction, retrieve a relevant memory, or repeat the rule in
its plan, but fail to carry that preference through to the final response or
workspace state. In other cases, the agent recognizes the preference but
treats it as secondary to completing the immediate task. These failures
illustrate why making past feedback available does not necessarily produce a
stable behavioral change. Appendix~\ref{sec:appendix:access-stores} provides
an anonymized example across the different representations.

These findings motivate turning user corrections into actionable
requirements that guide future execution and support explicit
checking when possible.
\begin{figure*}[t]
\centering
\includegraphics[width=\linewidth]{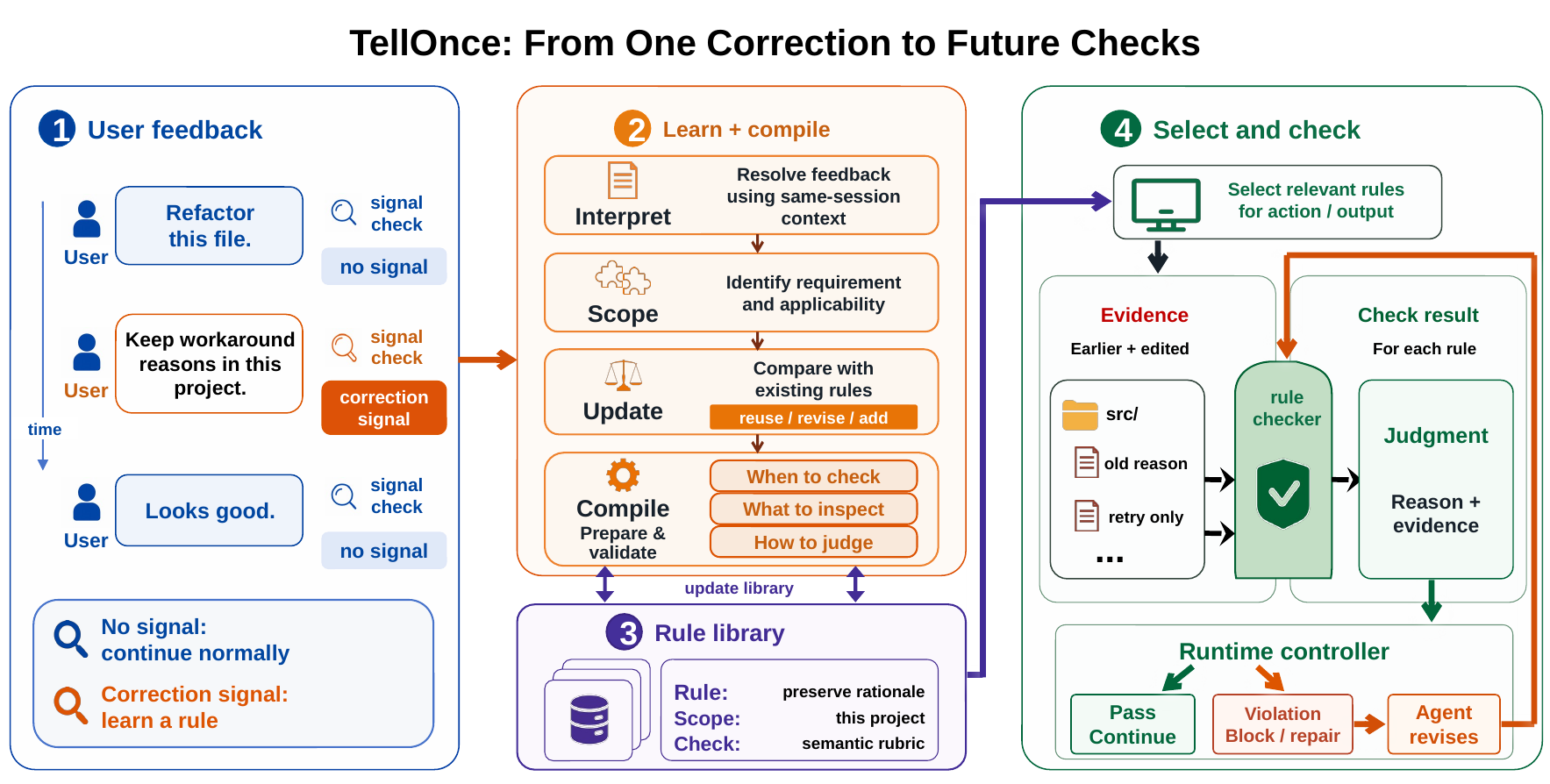}
\caption{Overview of \sysname{}. Left: user feedback and correction
detection. Center: rule learning, check compilation, and storage.
Right: rule selection, runtime checking, and intervention through the
\sysname{} execution layer.}
\label{fig:trace-pipeline}
\end{figure*}

\section{TellOnce: Turning Corrections into Checks}
\label{sec:trace}

\sysname{} converts user corrections into reusable behavioral rules and
runtime checks without updating the coding agent's model parameters.
A background learner identifies the requirement expressed by a correction,
determines its scope, reconciles it with existing rules, and prepares a
corresponding check when possible. In later tasks, relevant rules guide the
agent, while the \sysname{} execution layer connects these rules and checks
to the agent's actual execution. It collects the evidence needed for checking
and uses check results to block prohibited actions or request revision
(Figure~\ref{fig:trace-pipeline}). Detailed implementation and reproducibility information is provided in Appendix~\ref{sec:appendix:trace-details}.

\subsection{Learning and Maintaining Scoped Rules}
\label{sec:trace:lifecycle}

When the user provides feedback, the learner first examines
the current message for a potentially reusable correction.
If one is detected, it consults recent conversation history
to resolve references and interpret the intended behavior,
while grounding any learned requirement in the user's own
words. The learner then determines whether the requirement
should persist and where it applies, such as across tasks
or within a particular project. Task-specific instructions are not stored as lasting
rules, and feedback with unresolved meaning or scope
does not become an active rule.

For each reusable correction, the learner extracts the
requested behavior, scope, conditions, and exceptions,
then reconciles it with the existing rule library.
It may reuse, revise, replace, or add rules, while
avoiding redundant or conflicting entries. Updates
must be grounded in the user's words, and existing
requirements are removed only when explicitly retracted.

\subsection{Compiling Rules into Check Specifications}
\label{sec:trace:compiled}

To make learned rules usable in future tasks, the compiler
assigns each rule one of four outcomes. Simple, global requirements without conditions or exceptions
can receive programmatic checks when compliance can be
determined directly. Rules requiring semantic interpretation
instead use an LLM-based checker with relevant execution evidence. If no reliable
check can be validated but the rule itself is clear, it
remains available as behavioral guidance without automated
checking. If the rule's meaning remains ambiguous, it is
left inactive and is neither shown to the agent nor checked.

For rules that are checked, the compiler prepares a reusable
\emph{check specification} describing when the rule applies, at which stage
it should be checked, what evidence is required, and what counts as compliance
or violation. The check must preserve the meaning, scope, and exceptions of
the original rule rather than change the requirement to make it easier to
verify.

Before activation, the compiler generates examples of compliance,
violation, and non-applicability. A separate LLM call checks whether
the proposal preserves the rule's meaning and independently labels
the examples without seeing the expected labels. The check is
activated only if the labels agree and its execution matches the
expected results on every example. Failed proposals can be revised
using the validation feedback for up to three rounds; otherwise,
the rule remains guidance-only or inactive.

\subsection{Checking and Intervening in Later Tasks}
\label{sec:trace:runtime}

At the start of each user turn, the system selects
applicable rules from the latest available library
based on the task, the user's latest instruction,
and each rule's scope and conditions. The selected
rules, including their conditions and exceptions,
are provided to the agent as guidance, together with
an explicit reminder that current user instructions
and required output formats take precedence over
learned rules. The corresponding compiled checks
are made available during execution. Learning
proceeds separately in the background, so newly
learned rules become available in later turns once
they are ready.

When a selected rule is checked, the \sysname{} execution layer
supplies evidence from the relevant stage, such as a proposed
action, a produced artifact, or a candidate reply. Programmatic
checks evaluate explicit conditions, while LLM-based checks
apply the compiled semantic criteria. Each check returns one
of four outcomes: satisfied, violated, not applicable, or
unknown. Insufficient evidence yields unknown, which is not
treated as compliance.

The execution layer uses these judgments to guide intervention.
Depending on the execution stage, it can reject a proposed
action, provide feedback on a produced artifact, or request
revision of a candidate reply. Revised outputs can be checked
again, allowing the agent to correct violations before delivery.
To prevent indefinite repair loops, the system limits revision
requests and records violations that remain unresolved.

\begin{table*}[t]
\centering
\small
\setlength{\tabcolsep}{3.6pt}
\renewcommand{\arraystretch}{1.20}
\caption{Results on new tasks without additional user corrections.
Task scores follow native benchmark scales; preference violation rates
are percentages. Mean~$\pm$~standard deviation over three simulated users.}
\label{tab:tellonce-main-results}
\begin{tabularx}{\linewidth}{@{}p{0.22\linewidth}l
    *{4}{>{\centering\arraybackslash}X}
    >{\columncolor{TellOnceTint}[\tabcolsep][0pt]\centering\arraybackslash}X@{}}
\arrayrulecolor{TellOnceAccent}
\toprule
\textbf{Benchmark} & \textbf{Metric}
& \shortstack{No\\Memory} & Mem0 & PGMem & Zoro
& \textcolor{TellOnceAccent}{\textbf{\sysname{}}} \\
\midrule

\multicolumn{7}{@{}>{\columncolor{TellOnceBand}[0pt][0pt]}l@{}}{\hspace{5pt}\rule{0pt}{2.4ex}%
\textcolor{TellOnceAccent}{\textbf{GPT-5.6 Luna}}} \\
\multirow{2}{*}{ClawArena}
& Task score $\uparrow$ & 77.4\,$\pm$\,1.9 & 77.0\,$\pm$\,1.1 & 77.0\,$\pm$\,1.1 & 75.4\,$\pm$\,0.1 & 77.2\,$\pm$\,1.8 \\
& Violation $\downarrow$ & 99.8\,$\pm$\,0.4 & 83.1\,$\pm$\,4.8 & 73.6\,$\pm$\,8.5 & 99.3\,$\pm$\,0.1 & \textbf{34.7\,$\pm$\,7.5} \\
\addlinespace[2pt]
\multirow{2}{*}{MemoryArena}
& Task score $\uparrow$ & 57.5\,$\pm$\,2.6 & 57.8\,$\pm$\,1.0 & 55.6\,$\pm$\,1.1 & 57.1\,$\pm$\,2.0 & 56.3\,$\pm$\,4.1 \\
& Violation $\downarrow$ & 56.5\,$\pm$\,6.9 & 53.6\,$\pm$\,3.2 & 36.5\,$\pm$\,10.8 & 55.4\,$\pm$\,2.3 & \textbf{13.6\,$\pm$\,2.0} \\
\addlinespace[5pt]

\multicolumn{7}{@{}>{\columncolor{TellOnceBand}[0pt][0pt]}l@{}}{\hspace{5pt}\rule{0pt}{2.4ex}%
\textcolor{TellOnceAccent}{\textbf{GPT-5.6 Sol}}} \\
\multirow{2}{*}{ClawArena}
& Task score $\uparrow$ & 77.8\,$\pm$\,2.9 & 76.7\,$\pm$\,1.2 & 77.5\,$\pm$\,0.4 & 77.6\,$\pm$\,1.2 & 78.5\,$\pm$\,0.2 \\
& Violation $\downarrow$ & 100.0\,$\pm$\,0.0 & 93.0\,$\pm$\,6.9 & 81.5\,$\pm$\,10.5 & 100.0\,$\pm$\,0.0 & \textbf{39.2\,$\pm$\,4.1} \\
\addlinespace[2pt]
\multirow{2}{*}{MemoryArena}
& Task score $\uparrow$ & 93.3\,$\pm$\,1.1 & 92.9\,$\pm$\,0.9 & 91.4\,$\pm$\,1.8 & 91.2\,$\pm$\,0.8 & 91.7\,$\pm$\,1.9 \\
& Violation $\downarrow$ & 56.2\,$\pm$\,0.8 & 47.5\,$\pm$\,7.6 & 15.6\,$\pm$\,1.8 & 57.8\,$\pm$\,3.9 & \textbf{12.6\,$\pm$\,6.4} \\
\addlinespace[5pt]

\multicolumn{7}{@{}>{\columncolor{TellOnceBand}[0pt][0pt]}l@{}}{\hspace{5pt}\rule{0pt}{2.4ex}%
\textcolor{TellOnceAccent}{\textbf{Claude Haiku 4.5}}} \\
\multirow{2}{*}{ClawArena}
& Task score $\uparrow$ & 74.0\,$\pm$\,1.6 & 72.7\,$\pm$\,1.7 & 67.7\,$\pm$\,1.1 & 72.9\,$\pm$\,0.9 &72.9\,$\pm$\,0.9 \\
& Violation $\downarrow$ & 98.8\,$\pm$\,0.6 & 96.1\,$\pm$\,2.3 & 88.1\,$\pm$\,10.9 & 99.0\,$\pm$\,0.3 & \textbf{68.8\,$\pm$\,5.7} \\
\addlinespace[2pt]
\multirow{2}{*}{MemoryArena}
& Task score $\uparrow$ & 53.7\,$\pm$\,2.5 & 54.6\,$\pm$\,3.6 & 48.0\,$\pm$\,0.4 & 39.8\,$\pm$\,6.2 & 53.4\,$\pm$\,1.6 \\
& Violation $\downarrow$ & 37.8\,$\pm$\,2.1 & 33.8\,$\pm$\,0.5 & 32.7\,$\pm$\,3.5 & 33.4\,$\pm$\,2.4 & \textbf{14.5\,$\pm$\,1.3} \\
\bottomrule
\end{tabularx}
\arrayrulecolor{black}
\end{table*}

\section{Experiments}
\label{sec:experiments}

\subsection{Experimental Setup}
\label{sec:experiments:setup}

\textbf{Benchmarks and agents.}
Our main evaluation uses ClawArena~\citep{ji2026clawarenabenchmarkingaiagents}
and the travel-planning tasks from
MemoryArena~\citep{he2026memoryarenabenchmarkingagentmemory}.
ClawArena contains 8 learning and 16 test scenarios; MemoryArena contains
30 learning and 60 test tasks. We split complete tasks rather than turns
within a task, preserving each task's original input sequence and context.
We evaluate all methods with three task-solving models: GPT-5.6 Luna,
GPT-5.6 Sol, and Claude Haiku 4.5.

\textbf{Simulated users and learning.}
Following prior work on simulated interaction for
personalization~\citep{ma2026syntheticinteractiondatascalable}, we use
three user profiles specifying preferences, their conditions, and exceptions.
During learning, the simulator uses the profile and interaction history
to decide whether the agent's response needs correction and generates
natural-language feedback. Each memory-based method starts with an empty
store for each model, user, and benchmark. Methods share the same learning
tasks, task order, and feedback policy, with at most one additional
correction per original benchmark question. The feedback responds to each
method's actual behavior. Agents receive full feedback, but only reusable preferences
update long-term memory. Hidden profiles and simulator labels
are not shared. Profile construction and
feedback generation are described in Appendix~\ref{sec:simulated-users}.

\textbf{Evaluation without further corrections.}
After learning, each method's long-term state is frozen.
Test tasks provide only the benchmark's original inputs,
without additional user corrections or answers to
agent-initiated questions. Agents may perform tool calls
and internal checks before producing one user-facing
response per original question. Execution state is
preserved within each task, but not across tasks.
Full protocols are provided in Appendix \ref{sec:appendix:experimental-details}.

\textbf{Comparison methods.}
We compare \sysname{} with \textsc{No Memory},
\textsc{Mem0}~\citep{mem0},
\textsc{PGMem}~\citep{choi2026pgmem}, and
\textsc{Zoro}~\citep{ma2026zoro}.
Each method uses its own memory-update, retrieval, and execution mechanisms.
\textsc{No Memory} retains normal task context, tool access, and self-checking
capabilities, but carries no long-term memory between tasks.

\textbf{Metrics.}
We report \emph{task score} and \emph{preference violation rate}.
Task score follows each benchmark's native scoring procedure.
Preference violation rate is the percentage of original evaluation units
(474 ClawArena rounds or 415 MemoryArena questions per user)
with at least one confirmed violation of an applicable user preference
in the delivered response or artifact. All units remain in the denominator,
including those with undetermined or inapplicable preference judgments.
Preference scoring uses the user profile, task context, and actual
deliveries, without access to method identities, stored memories,
or internal checking and repair results. Evaluation scores are not
returned to the agent. Further details are provided in
Appendix~\ref{sec:appendix:experimental-details}.

\subsection{Main Results}
\label{sec:experiments:main-results}

\textbf{Preference following.}
Table~\ref{tab:tellonce-main-results} shows that \sysname{} achieves
the lowest preference violation rate across all three models on both
benchmarks. PGMem achieves the lowest preference violation rate among
the baselines in all six settings.
On ClawArena, \sysname{} reduces violations relative to PGMem by
19.3--42.3 percentage points, including reductions from 73.6\% to
34.7\% for Luna and from 81.5\% to 39.2\% for Sol.
On MemoryArena, \sysname{} achieves violation rates of 12.6--14.5\%
across the three models, with improvements over PGMem ranging from
3.0 to 22.9 percentage points. These results demonstrate that
\sysname{} more reliably carries learned user requirements into later
tasks without requiring repeated corrections.

\textbf{Task performance.}
The improvements in preference following do not consistently translate
into higher native task scores. Across all six settings, \sysname{} achieves task scores
within approximately 1.6 percentage points of the
strongest baseline. Overall, the results show substantial improvements in
following previously learned user requirements, while task-completion
performance remains largely comparable, with some setting-specific
trade-offs.

\subsection{Ablation Study}
\label{sec:experiments:ablation}

We study three parts of how \sysname{} uses learned rules in later tasks:
selecting rules relevant to the current task, checking whether the agent
follows those rules, and revising the final output when a violation is found.
All conditions use the same frozen rule library.

\begin{table*}[t]
\centering
\small
\setlength{\tabcolsep}{3.6pt}
\renewcommand{\arraystretch}{1.20}

\caption{Ablation of how learned rules are used at test time.
All conditions use the same frozen rule library.
\textsc{Before Final Repair} denotes the saved candidate immediately before
the final delivery-stage repair in the corresponding full \sysname{} run.
Values are percentages, mean~$\pm$~standard deviation over three simulated users.}
\label{tab:tellonce-ablation}

\begin{tabularx}{\linewidth}{
    @{}p{0.22\linewidth}l
    *{3}{>{\centering\arraybackslash}X}
    >{\columncolor{TellOnceTint}[\tabcolsep][0pt]
      \centering\arraybackslash}X@{}
}
\arrayrulecolor{TellOnceAccent}

\toprule
\textbf{Benchmark} & \textbf{Metric}
& \shortstack{All Rules\\as Text}
& \shortstack{Selected Rules\\as Text}
& \shortstack{Before Final\\Repair}
& \textcolor{TellOnceAccent}{\textbf{\sysname{}}} \\
\midrule

\multicolumn{6}{
    @{}>{\columncolor{TellOnceBand}[0pt][0pt]}l@{}
}{
    \hspace{5pt}\rule{0pt}{2.4ex}
    \textcolor{TellOnceAccent}{\textbf{GPT-5.6 Luna}}
} \\

\multirow{2}{*}{ClawArena}
& Task score $\uparrow$
& 76.3\,$\pm$\,1.4 & 76.7\,$\pm$\,0.8 & 77.3\,$\pm$\,1.7 & 77.2\,$\pm$\,1.8 \\
& Violation $\downarrow$
& 54.6\,$\pm$\,18.8 & 49.8\,$\pm$\,16.0 & 36.8\,$\pm$\,6.6 & \textbf{34.7\,$\pm$\,7.5} \\

\addlinespace[2pt]

\multirow{2}{*}{MemoryArena}
& Task score $\uparrow$
& 56.7\,$\pm$\,0.8 & 56.3\,$\pm$\,0.7 & 55.2\,$\pm$\,4.1 & 56.3\,$\pm$\,4.1 \\
& Violation $\downarrow$
& 16.8\,$\pm$\,3.3 & 14.3\,$\pm$\,2.3 & 13.8\,$\pm$\,2.2 & \textbf{13.6\,$\pm$\,2.0} \\

\addlinespace[5pt]

\multicolumn{6}{
    @{}>{\columncolor{TellOnceBand}[0pt][0pt]}l@{}
}{
    \hspace{5pt}\rule{0pt}{2.4ex}
    \textcolor{TellOnceAccent}{\textbf{GPT-5.6 Sol}}
} \\

\multirow{2}{*}{ClawArena}
& Task score $\uparrow$
& 78.1\,$\pm$\,1.2 & 78.2\,$\pm$\,1.5 & 78.4\,$\pm$\,0.3 & 78.5\,$\pm$\,0.2 \\
& Violation $\downarrow$
& 58.7\,$\pm$\,14.4 & 55.5\,$\pm$\,11.0 & 44.1\,$\pm$\,6.8 & \textbf{39.2\,$\pm$\,4.1} \\

\addlinespace[2pt]

\multirow{2}{*}{MemoryArena}
& Task score $\uparrow$
& 91.6\,$\pm$\,0.6 & 91.0\,$\pm$\,0.7 & 91.6\,$\pm$\,2.1 & 91.7\,$\pm$\,1.9 \\
& Violation $\downarrow$
& 16.9\,$\pm$\,4.0 & 14.1\,$\pm$\,6.4 & 13.6\,$\pm$\,6.8 & \textbf{12.6\,$\pm$\,6.4} \\

\addlinespace[5pt]

\multicolumn{6}{
    @{}>{\columncolor{TellOnceBand}[0pt][0pt]}l@{}
}{
    \hspace{5pt}\rule{0pt}{2.4ex}
    \textcolor{TellOnceAccent}{\textbf{Claude Haiku 4.5}}
} \\

\multirow{2}{*}{ClawArena}
& Task score $\uparrow$
& 73.5\,$\pm$\,2.0 & 72.9\,$\pm$\,1.0 & 72.8\,$\pm$\,0.9 & 72.9\,$\pm$\,0.9 \\
& Violation $\downarrow$
& 86.3\,$\pm$\,8.9 & 84.7\,$\pm$\,2.7 & 74.0\,$\pm$\,16.0 & \textbf{68.8\,$\pm$\,5.7} \\

\addlinespace[2pt]

\multirow{2}{*}{MemoryArena}
& Task score $\uparrow$
& 54.1\,$\pm$\,3.2 & 53.7\,$\pm$\,2.3 & 52.3\,$\pm$\,1.9 & 53.4\,$\pm$\,1.6 \\
& Violation $\downarrow$
& 18.2\,$\pm$\,2.8 & 17.2\,$\pm$\,1.1 & 16.0\,$\pm$\,1.7 & \textbf{14.5\,$\pm$\,1.3} \\

\bottomrule
\end{tabularx}
\end{table*}

\textbf{Does selecting relevant rules help?}
Comparing \textsc{All Rules as Text} with \textsc{Selected Rules as Text},
we find that rule selection reduces preference violations in all six
settings by 1.0--4.8 percentage points, while task scores differ by
at most 0.6 points. This suggests that focusing the agent on relevant
rules helps, but selection alone provides relatively modest gains.

\textbf{Does checking and intervention improve compliance?}
Compared with providing selected rules as text, full \sysname{}
substantially reduces violations on ClawArena: from 49.8\% to 34.7\%
for Luna, 55.5\% to 39.2\% for Sol, and 84.7\% to 68.8\% for Haiku.
On MemoryArena, selected rule guidance already achieves relatively
low violation rates, and the full system provides smaller additional
improvements of 0.7--2.7 percentage points. These results suggest
that making relevant rules available does not always translate into
following them during execution, and that checking and intervention
provide additional benefits, particularly on ClawArena.

\textbf{How much does final repair contribute?}
We compare the saved candidate immediately before final delivery-stage
repair with the final output from the same \sysname{} run.
Final repair further reduces violations by 2.1--5.2 percentage points
on ClawArena and 0.2--1.5 points on MemoryArena, with generally
similar task scores. Since earlier execution-stage checks may already
have influenced the saved candidate, this comparison measures the
additional benefit of final repair rather than the effect of all
checking.

\subsection{Reusing Task-Improvement Feedback}
\label{sec:experiments:interactive-tasks}

\begin{wrapfigure}{R}{0.5\textwidth}
    \centering
    \includegraphics[width=\linewidth]{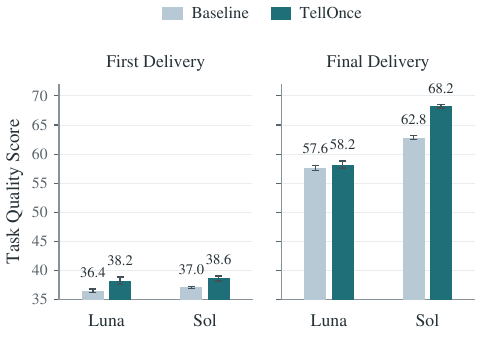}
    \caption{Reusing task-improvement feedback on SWE-Together.
    Bars show mean first- and final-delivery task quality over three
    independent runs; error bars indicate one sample standard deviation.
    Higher is better. }
    \label{fig:task-improvement-results}
\end{wrapfigure}

We next ask whether \sysname{} can learn from feedback whose goal is to improve
task completion, rather than primarily encode a user-specific preference.
We use a general simulated user that provides task-relevant information,
clarifies requirements, and points out problems in the current solution.
We then evaluate whether experience accumulated from such interactions improves
performance on later tasks.

\textbf{SWE-Together~\citep{wu2026swe}.}
For each of GPT-5.6 Luna and GPT-5.6 Sol, \sysname{} learns sequentially from
60 tasks and is then evaluated on 49 paired test tasks with its long-term state
frozen. We compare against the same solver without cross-task memory.
SWE-Together preserves normal multi-turn interaction within each test task, so
we evaluate both the first user-facing delivery and the final delivery after
the full interaction. The first delivery provides the cleaner measure of
cross-task reuse because it is produced before the user supplies new feedback
for the current task.

%
\textbf{Results.}
Figure~\ref{fig:task-improvement-results} shows that \sysname{}
improves both first- and final-delivery task quality on SWE-Together.
Before receiving additional task-specific feedback, first-delivery
scores increase from 36.4 to 38.2 for Luna and from 37.0 to 38.6
for Sol. After the full within-task interaction, \sysname{} achieves
58.2 versus 57.6 for Luna and 68.2 versus 62.8 for Sol.
The improvement remains modest for Luna at the final delivery
(+0.6 points), but becomes substantially larger for Sol (+5.4 points).
These results show that the benefits of reusing past feedback extend
beyond the initial response and can persist throughout subsequent
task interaction.

Full task selection, interaction protocols, scoring procedures, and execution
details are provided in
Appendix~\ref{sec:appendix:experimental-details}.

\section{Conclusion}

We identify an access--compliance gap: agents may
repeat previously corrected behavior even when
relevant feedback is available. To address this,
we introduce \sysname{}, which converts user
corrections into persistent, scoped, and revisable
behavioral rules, supported by execution-time checks
when reliable verification is possible. Across
ClawArena and MemoryArena, \sysname{} substantially
reduces preference violations while maintaining
broadly comparable task performance. SWE-Together
experiments further show that feedback aimed at
improving task completion can benefit later tasks.
\section*{Limitations}

Our evaluation primarily relies on simulated users to provide feedback
during interaction. Although we separately validate the simulation
procedure against real user correction histories, simulated feedback
cannot fully capture the diversity and unpredictability of long-term
human--agent collaboration. Our experiments also cover a fixed set
of agent models and benchmark tasks, so broader generalization to
other environments and user populations remains to be studied.

\sysname{} relies on correctly interpreting what a user wants the agent
to learn and when that requirement should apply. Ambiguous or
context-dependent corrections may be difficult to interpret reliably.
Moreover, not every learned requirement can be checked automatically
from available execution evidence. In such cases, \sysname{} retains
the requirement as behavioral guidance, while semantic checking
may also introduce judgment errors.

Finally, although \sysname{} supports revising previously learned
rules, our evaluation of rule updates is primarily based on controlled
component diagnostics rather than extended interactions in which
user preferences repeatedly change. Evaluating such long-term
adaptation in real deployments remains an important direction.
Rule learning, selection, and checking also introduce additional
computation, which is not explicitly matched across methods in
our end-to-end comparisons.

\subsection*{AI use statement}
Generative AI tools were used to assist with drafting and
revising portions of the manuscript, improving clarity,
and preparing LaTeX formatting. The authors take full
responsibility for the scientific claims, experimental
results, and final content of this paper.

\bibliography{references}

\clearpage
\appendix
\begin{table*}[t]
    \centering
    \small
    \caption{Preference stores used in the access--compliance diagnostic.
    Counts refer to the stores available to each condition rather than to the
    29 held-out preference checks.}
    \label{tab:access-store-summary}
    \begin{tabular}{p{0.22\linewidth}p{0.14\linewidth}p{0.54\linewidth}}
        \toprule
        Store & Size & Representation and use \\
        \midrule

        Prompt rule store &
        29 rules &
        Hand-curated natural-language rules representing the target
        preferences. \textsc{All Rules} receives the complete set in context,
        while \textsc{Relevant Rules} receives only the rules selected as
        applicable to the current task. These conditions test how well the
        agent follows preferences expressed directly as natural-language
        guidance. \\

        Operational rule library &
        47 entries &
        Structured rules constructed by \sysname{} from the 123 training
        correction-conflict records. The library contains 37 atomic entries,
        7 refinements, and 3 composite entries. These entries cover the same
        canonical preferences as the 29-rule prompt store, but preserve
        additional structure for scope and applicability; three composite
        preferences are decomposed into seven refinement entries for more
        specific use. For the \textsc{Compiled Rules} condition, applicable
        entries additionally include explicit criteria describing how the
        corresponding behavior should be assessed. \\

        Mem0 store &
        121 records &
        A local Mem0 collection constructed from the same 123 training
        correction-context records, excluding all 19 held-out tasks. Mem0
        stores 121 records after its ingestion procedure merges or removes
        semantically overlapping information. Embeddings use OpenAI
        \texttt{text-embedding-3-small}, with a local Chroma vector store.
        At evaluation time, up to five memories are retrieved for the task
        prompt and inserted into the agent context. \\

        \bottomrule
    \end{tabular}
\end{table*}

\section{Access-Compliance Diagnostic Details}
\label{sec:appendix:access-stores}

This appendix supplements Section~\ref{sec:access-compliance} with the
construction of the diagnostic dataset, the held-out task selection procedure,
and the preference representations used by each condition.

\subsection{Data Collection, Held-Out Task Selection, and Annotation}
\label{sec:appendix:access-stores:collection}

The diagnostic uses correction cases from a single user's historical
coding-agent interactions. We use a single-user corpus deliberately so that
the diagnostic measures whether an agent can act on an available preference,
without introducing variation from different users' preferences. The source
corpus contains 32 long-context transcripts collected over approximately two
months of daily use, from which we extracted 142 correction-conflict records:
cases in which a user correction could be paired with an agent behavior that
violated it. All transcripts were anonymized before annotation by removing
project-specific names, file paths, and identifying details while preserving
the structure of the interaction.

We select the 19 held-out evaluation tasks from these records using three
filters:
\begin{enumerate}
    \item \textbf{Repetition.} The underlying preference must reflect a
    recurring rather than one-off source of friction.
    \item \textbf{Deduplication.} Near-duplicate cases expressing the same
    preference are represented by a single held-out task.
    \item \textbf{Self-contained context.} The task must retain enough context
    to be attempted without requiring access to the original transcript.
\end{enumerate}

The 19 tasks contain 29 manually annotated preference checks. The held-out
tasks are excluded from all store construction: both the operational rule
library and the Mem0 store are built only from the remaining 123
correction-conflict records. Thus, the target preference for a held-out task
is never learned from that task itself.

\paragraph{Scope of the diagnostic.}
The diagnostic is intended to isolate the access--compliance question rather
than estimate population-level preference-following performance. Its absolute
compliance rates therefore reflect the preference distribution of this
single-user corpus. The main experiments in Section~\ref{sec:experiments}
complement this diagnostic by evaluating correction reuse across three
distinct user profiles on ClawArena and MemoryArena; see
Appendix~\ref{sec:simulated-users}.

\subsection{Preference Stores}
\label{sec:appendix:access-stores:stores}

Table~\ref{tab:access-store-summary} describes the preference representations
used in the diagnostic conditions, and
Table~\ref{tab:access-store-example} illustrates how the same underlying
correction is represented across them. The key distinction is not simply
whether the preference is available to the agent, but what information the
representation provides about when the preference applies and how its
satisfaction should be assessed in a later task.

\begin{table*}[t]
    \centering
    \small
    \caption{An anonymized example showing how the same underlying correction
    is represented across diagnostic conditions. The example is paraphrased;
    raw transcript text is not reproduced.}
    \label{tab:access-store-example}
    \begin{tabular}{p{0.19\linewidth}p{0.33\linewidth}p{0.38\linewidth}}
        \toprule
        Component & Example content & Use at evaluation time \\
        \midrule

        Source correction &
        ``You left debug files again. Clean them before stopping.'' &
        The historical correction motivates the preference, but the correction
        itself is not included in the held-out task prompt. \\

        Prompt rule &
        Before stopping, remove temporary or debug files introduced during the
        task. &
        The preference is provided directly as natural-language guidance,
        either as part of the full rule set or after task-specific selection. \\

        Mem0 memory &
        The user prefers cleanup of debug artifacts before finalizing coding
        tasks. &
        Mem0 may retrieve this representation for a related task and place it
        in context. Retrieval makes the preference available, but the stored
        text does not itself specify an explicit criterion for evaluating the
        final workspace. \\

        Operational rule &
        Applies when the task may create workspace artifacts; evaluate the
        final workspace for temporary or debug files before completion. &
        The structured representation preserves applicability information and
        associates the preference with an explicit criterion that can be
        evaluated against the resulting workspace. \\

        \bottomrule
    \end{tabular}
\end{table*}
\section{TellOnce: Reproducibility Details}
\label{sec:appendix:trace-details}

\sysname{} organizes reusable corrections around a common lifecycle. It first
turns user feedback into durable behavioral rules, then prepares a reusable
checking procedure when possible, and finally uses those rules and checks in
later tasks. The system combines a rule-learning and checking core with an
execution layer for coding-agent environments. The core maintains rules,
prepares check specifications, and returns structured judgments on supplied
evidence. The \sysname{} \emph{execution layer} connects these capabilities to
the coding agent by providing rule guidance, collecting execution evidence,
triggering checks, and using their results at the control points exposed by
the underlying platform.

The same rule and checking abstractions are used across platforms, while the
available hooks determine where guidance, checking, and intervention can occur.
This appendix describes rule acquisition, check compilation, runtime checking,
and the native integrations for Claude Code, Codex, and Copilot CLI.

\subsection{Rule Acquisition and Library Maintenance}
\label{sec:appendix:trace-details:lifecycle}

\paragraph{Admission and interpretation.}
The correction learner records behavioral requirements that should remain
useful in future tasks, including user preferences, recurring sources of
friction, and mistakes the user wants the agent to avoid. Task answers and
instructions that apply only to the current task are not automatically stored
as durable rules.

For each user message, signal detection first examines
the message alone, erring toward identifying a possible
learning signal when uncertain. If no signal is found,
the rule library remains unchanged. Otherwise, the learner
reads recent same-session context to clarify what the
correction refers to, without treating that context as an
independent source of requirements.

A separate interpretation step identifies what the user
rejects, what remains acceptable, and what reusable
behavior the correction may imply, supported by exact
quotations from the user's message. This interpretation
serves only as a hypothesis. Rule resolution then checks
it against the user's original words and the existing rule
library to determine the appropriate update. If a reference
remains unclear, the system retrieves up to five exact
excerpts from the same session and retries resolution once.
For example, ``Refactor this file'' is a task request, while ``for this
refactor only, add no dependencies'' is a temporary constraint. Neither
establishes a lasting preference. In contrast, an explicit statement such as
``in this project, do not add new dependencies'' can establish a project-level
rule without requiring repeated complaints. If the requested behavior or its
scope cannot be determined reliably, no durable rule is created from that
feedback.

\paragraph{Updating the rule library.}
A new correction is compared with the existing rule library rather than
appended as an independent memory by default. If an existing rule already
captures the requirement, it can be reused. If the user changes the behavior,
scope, condition, or exception, the corresponding rule can be revised or
superseded. A new rule is added only when the requirement is not already
represented.

Independent requirements can be stored separately, while an exception remains
attached to the rule it modifies. Each stored rule retains the user requirement,
its scope, conditions, exceptions, and supporting feedback so that later
corrections can update the same rule rather than accumulate duplicate or
inconsistent memories.

Each proposed update must be supported by the user's original
words. Existing requirements can be removed only when explicitly
retracted, and new rules are added only when the requirement is
not already covered by an existing rule. Proposals that violate these constraints
are returned once for revision and discarded if they
remain invalid.

\paragraph{Temporary exceptions and lasting updates.}
A new instruction does not always imply that an existing durable rule should
change. Suppose the user previously asked for concise answers. A later request
such as ``be detailed this time'' changes only the current task and leaves the
stored preference unchanged. In contrast, ``from now on, be more detailed''
expresses a lasting change and is processed as an update to the existing rule.

The learner therefore distinguishes temporary exceptions from durable changes
based on the user's current instruction and its intended duration. Temporary
exceptions affect only the current interaction; lasting changes update the
stored rule through the same rule-resolution process described above.

\paragraph{Library versions and adoption.}
A complete library version associates each rule revision
with its compilation outcome. Validated checks are reused
when the corresponding rule and compiler configuration
remain unchanged. Rules that remain guidance-only or
inactive are reconsidered for compilation during the next
two library updates and once more at the end of training.
Guidance-only rules are presented to the agent but are
never checked, whereas inactive rules are neither
presented nor checked.

At the beginning of a user turn, execution adopts a complete rule-library
version and keeps that version fixed for the turn. Rule learning and check
preparation can continue separately in the background. Once a new version is
fully prepared, it becomes available at a later user-input boundary rather than
changing the rule set in the middle of an ongoing turn. This keeps the rule
guidance shown to the agent and the corresponding check specifications
consistent within each execution turn.

\subsection{Check Compilation and Validation}
\label{sec:appendix:trace-details:validation}

\paragraph{Choosing how a rule is checked.}
Once a durable rule has been learned, the compiler determines
how that rule can be used in later tasks. There are four outcomes.

First, global rules without conditions or exceptions can use
programmatic checks when their requirements are directly
computable. The current implementation supports three types:
requiring fields in a JSON reply, enforcing an explicit word
limit, and comparing a field of an action, artifact, or reply
with a required literal value. These checks do not require
an additional model call.

Second, rules that require semantic interpretation use an
LLM-based check. For these rules, the compiler prepares a
semantic specification describing when the rule applies,
what behavior is required, what constitutes compliance or
violation, and what evidence is needed at each checking stage.

Third, if a rule is clear enough to guide the agent but no
reliable automated check can be validated, it remains available
as behavioral guidance without an executable check.

Fourth, if the meaning of a rule remains ambiguous, the rule
is left inactive: it is neither presented to the agent nor
checked.

For executable checks, the specification must preserve the meaning, scope,
conditions, and exceptions of the learned rule. The compiler does not simplify
a requirement merely to make it easier to verify. A check also cannot depend
on evidence that does not yet exist: for example, a pre-action check cannot
require an artifact that will only be produced after the action executes.

\paragraph{Validation and fallback.}
Before an executable check is activated, the compiler
generates 3--9 validation cases covering compliance,
violation, and non-applicability. Each selected checking
stage includes both compliant and violating cases, and
the complete set includes at least one non-applicable case.

A separate reviewer receives the rule, proposed check,
and validation cases, but not the compiler's expected
labels or rationale. It assesses whether the proposal
faithfully represents the rule, whether the requirement
is sufficiently clear, and independently labels each case.
Any disagreement with the compiler's expected labels
rejects the proposal.

If the review succeeds, every case is executed through
the actual checking procedure. Programmatic cases run
locally, while semantic cases invoke the model checker.
All resulting judgments must match the expected labels
before the check can be activated.

If either review or execution validation fails, the compiler
uses the rejection reason to revise its proposal, for up to
three rounds. A rule that still lacks a validated check is
retained as guidance if its meaning is clear, or left
inactive otherwise. Transport and output-format failures
may be retried once per model request, independently of
the proposal revision process.

\begin{table*}[t]
\centering\small
\caption{Illustrative example of a correction, its prepared semantic check,
and expected judgments on evidence from a later task.}
\label{tab:tellonce-example}
\label{tab:tellonce-rubric}
\label{sec:trace:example}
\begin{tabular}{@{}p{0.21\linewidth}p{0.73\linewidth}@{}}
\toprule
Component & Specification or expected behavior \\
\midrule
User feedback & ``In this project, preserve comments explaining why a workaround is needed.'' \\
Learned rule & Preserve the workaround's rationale when editing the identified project; equivalent rewording is allowed. \\
Applicability and stage & The edit affects a workaround explanation in that project. Use the artifact stage when the execution layer supplies the edited file. \\
Required evidence & Task/project context, resulting file, and earlier captured events containing the pre-edit explanation. \\
Decision criterion & The workaround's reason must remain understandable. Removing or materially changing that reason while retaining the workaround violates the rule; preserving the meaning satisfies it. \\
\midrule
Earlier observation & A recorded read contains ``Retry once because the legacy proxy closes idle connections.'' \\
Expected failure & The edited file retains the retry but removes its explanation. Return a failed judgment identifying the missing rationale. The execution layer can use this feedback to request a file revision; the write has already occurred. \\
Expected pass & The resulting comment says ``Legacy proxy drops idle connections; retry once.'' The same rationale remains. \\
Not applicable & The edit is outside the identified project or does not affect a workaround explanation. \\
Unknown & The earlier explanation is missing from the supplied evidence, so its preservation cannot be compared. \\
\bottomrule
\end{tabular}
\end{table*}

\subsection{Checking Evidence and Platform Integration}
\label{sec:appendix:trace-details:interfaces}

\paragraph{Selecting rules in later tasks.}
At each user-input boundary, the selector receives the task
description, the user's latest instruction, and all active
rules with their scopes and conditions, processed in batches
of up to 32. The initial selection considers applicability
to the entire response. Selected rules are provided to the
agent in full, including their types, scopes, conditions,
and exceptions. A fixed instruction clarifies that current
user requirements and submission formats take precedence
over learned rules, which must not be treated as factual
evidence about the task.

The execution layer can repeat rule selection for a specific
action, artifact, or candidate reply while keeping the same
rule-library version. In our experiments, the initial selection
was reused for actions, only the first produced file in each
turn triggered another selection, and unchanged files were
not checked again. These optimizations reduce repeated
selection and checking without modifying the stored rules.

\paragraph{Execution evidence and checking stages.}
The \sysname{} execution layer connects the checker to the coding agent's
actual execution. It constructs an observation containing the current task,
execution stage, and the evidence available at that point, such as a proposed
action, a produced artifact, a candidate reply, or previously recorded events.

The core supports three checking stages:
\begin{itemize}
    \item \texttt{pre\_action}: evidence about an operation before it executes;
    \item \texttt{artifact}: evidence about files or other artifacts after they
    are produced or modified;
    \item \texttt{delivery}: evidence about the candidate reply and relevant
    artifacts before the result is returned.
\end{itemize}

Programmatic checks evaluate explicit conditions directly.
An LLM-based check receives the learned rule, its compiled
specification, and only the evidence required by that
specification at the current stage. It also receives other
rules selected for the same reply, preventing content required
by one rule from being incorrectly judged as a violation of
another. Checks requiring the same evidence are grouped into
a single model call, with up to six rules evaluated separately.
For each rule, the checker returns a judgment, a reason, and
verbatim quotations from the supporting evidence.

\begin{table*}[t]
\centering\small
\caption{Core check results and their execution interpretation.}
\label{tab:tellonce-transitions}
\begin{tabular}{@{}p{0.20\linewidth}p{0.36\linewidth}p{0.37\linewidth}@{}}
\toprule
Result & Meaning & Execution interpretation \\
\midrule
\texttt{pass}
& The supplied evidence satisfies the check.
& No blocking action is triggered by this result. \\

\texttt{fail}
& The supplied evidence establishes a violation.
& Can provide a signal for blocking, revision, or feedback at an available
control point. \\

\texttt{not\_applicable}
& The rule or check does not apply at the current stage.
& No intervention is triggered for this check at this stage. \\

\texttt{unknown}
& The available evidence is insufficient for a reliable judgment.
& Does not itself trigger blocking or revision. \\

\texttt{error}
& The checking request cannot be completed or validated.
& Follows the error-handling behavior configured for the corresponding
integration. \\
\bottomrule
\end{tabular}
\end{table*}

An \texttt{unknown} result indicates insufficient evidence
for a reliable judgment. It does not independently trigger
blocking or repair, and a valid \texttt{unknown} result is not
retried. Before an action or after an artifact is produced,
the agent receives a non-blocking reminder; at delivery,
the uncertainty is recorded without preventing delivery.
Importantly, an \texttt{unknown} result cannot clear an
existing violation of the same rule on an unchanged candidate.
Only a subsequent \texttt{pass} or \texttt{not\_applicable}
judgment can clear that violation. Malformed checker responses
follow the bounded request-retry procedure.

\paragraph{Capturing execution evidence.}
The execution layer explicitly collects the evidence passed to the checker.
Action and artifact observations can include tool arguments, file paths,
content, and deletion status, while candidate replies and earlier events can
provide additional context.

Checks that compare an earlier and a later state require both states to be
present in the captured evidence. For example, a semantic comparison between
two file versions requires the relevant text from both versions; a filename
or hash alone is insufficient for that judgment. If required evidence is not
available, the checker can return \texttt{unknown}.

\paragraph{Platform adapters and triggers.}
The released plugins use each platform's native
rule-presentation and hook interfaces. By default, a
progressive index presents compact rule descriptions and
applicability information to the agent, retaining priority
rules where possible and rotating others when the index
exceeds its capacity. Semantic and keyword retrieval are
also available as alternative selection methods.
Table~\ref{tab:tellonce-host-interfaces} summarizes the
registered hooks and control behavior of these plugins,
which differ from the experimental execution layer.

\begin{table*}[htbp]
\centering\small
\setlength{\tabcolsep}{3pt}
\renewcommand{\arraystretch}{1.08}
\caption{Native plugin integration: rule presentation, checking triggers, and
available control behavior.}
\label{tab:tellonce-host-interfaces}
\begin{tabular}{@{}p{0.15\linewidth}p{0.24\linewidth}p{0.28\linewidth}p{0.25\linewidth}@{}}
\toprule
Platform & Guidance trigger & Checking input and trigger & Control behavior \\
\midrule

\raggedright Claude Code
& \raggedright \texttt{UserPromptSubmit}: rule guidance and prior alerts.
& \raggedright \texttt{Stop}: reply/transcript checks; optional semantic
judging.
& \raggedright Optional deterministic stop blocking; semantic judgments can
produce alerts for later guidance.
\tabularnewline

\raggedright Codex
& \raggedright \texttt{UserPromptSubmit}: rule guidance and prior alerts.
& \raggedright \texttt{PostToolUse}: tool-related evidence; an optional wrapper
also checks subprocess output.
& \raggedright Post-tool feedback or block signals when enabled; wrapper checks
are recorded after execution.
\tabularnewline

\raggedright Copilot CLI
& \raggedright \texttt{SessionStart}: rule guidance and prior alerts.
& \raggedright \texttt{Stop}: reply/transcript checks; optional semantic
judging.
& \raggedright Optional deterministic stop blocking through the platform
adapter; semantic judgments can produce alerts.
\tabularnewline

\bottomrule
\end{tabular}
\end{table*}

\paragraph{Experimental execution layer.}
Our experiments use a separate execution layer for Codex
and Claude Code, with common integration points before
tool calls, after tool calls, and before final delivery.
Selected learned rules are included directly in the
user turn rather than through the native plugins'
session-start or prompt-submission hooks. Before a tool
call, the execution layer can reject the proposed action;
after a tool call, it can provide feedback but cannot
undo a completed operation; before delivery, it can
request revision of the candidate reply and recheck
the result.

\paragraph{Checking and intervention paths.}
Three checking paths serve different roles in the current integrations.

\emph{Core-compiled checks} evaluate requirements learned from user feedback
using the check specifications prepared during compilation.

\emph{Native programmatic checks} evaluate explicit predicates registered in
the platform integration. They provide direct checks at the control points
supported by the corresponding coding-agent environment.

\emph{Semantic advisory checks} evaluate configured rules against reply
content and can produce alerts that are later returned to the agent as
guidance.

At a platform control point that supports blocking, a detected violation can
produce a block signal or feedback requesting revision. A later check can then
evaluate the revised evidence. Other checks can instead record feedback for
subsequent interaction.

\paragraph{Repair limits and appeals.}
The execution layer maintains intervention state across
checks within each user turn. Violations identified in the
same candidate reply are combined into a single revision
request. Up to three revision requests are allowed per user
input, shared across all rules and checking stages. Revised
outputs can be checked again, and any unresolved violations
are recorded when the repair budget is exhausted.

The agent can also appeal a disputed judgment once per
delivery by providing supporting evidence. A separate review
call determines whether to uphold or overturn the judgment.
An overturned judgment applies only to the unchanged
candidate output and does not modify the stored rules.

\subsection{Model Calls and Reproducibility Configuration}
\label{sec:appendix:trace-details:configuration}

This section reports the model and runtime configurations used in the
experiments in Section~5. The \sysname{} core is model-agnostic at the interface
level; the reported experiments use the fixed model assignments below.

\paragraph{Experimental model configuration.}
All auxiliary model calls in our experiments use GPT-6 Luna.
These include rule learning, check compilation and validation,
runtime checking, appeal review, simulated-user feedback
generation, preference scoring, and the memory-management
calls of Mem0, PGMem, and Zoro. These calls use low reasoning
effort. The maximum output length is 16,384 tokens
per call, or 8,192 tokens for each simulated-user step.
No temperature or other sampling parameters are specified.

The task-solving agents use GPT-5.6 Luna through Codex
with medium reasoning effort, GPT-5.6 Sol through Codex
with low reasoning effort, and Claude Haiku 4.5 through
Claude Code with default settings.

\begin{table*}[t]
\centering
\footnotesize
\setlength{\tabcolsep}{4pt}
\renewcommand{\arraystretch}{1.15}
\caption{Model-call contracts for rule learning, checking,
and evaluation. Conditional calls are invoked only when needed.}
\label{tab:tellonce-calls}
\begin{tabular}{@{}p{0.19\textwidth}p{0.34\textwidth}p{0.40\textwidth}@{}}
\toprule
Role & Input & Output \\
\midrule

Signal detection
& Current user message
& Whether a potential learning signal is present. \\

Correction interpretation
& User message and recent same-session context
& Rejected and retained behavior, possible reusable
requirements, and exact user quotations. \\

Rule resolution
& User message, interpretation, and existing rules
& Library change (keep, revise, replace, split, add,
defer, or reject), with supporting quotations. \\

Learning-time candidate selection
& User message, context, and batches of existing rules
& Candidate rules for resolution when the library
requires preliminary selection. \\

Context lookup
& Unclear correction and available session history
& Up to five relevant excerpts quoted exactly
from the session. \\

Check proposal
& Complete rule revision
& Check type, stages, specification, labeled
validation cases, and rationale. \\

Compiler review
& Rule, proposed check, and validation cases
without proposed labels or rationale
& Fidelity and clarity judgments, independent
case labels, and review feedback. \\

Example validation
& Proposed check and labeled validation cases
& Actual check results and their agreement
with expected labels. \\

Runtime rule selection
& Task, latest instruction, execution context,
and active rules
& Applicable rules for the current execution context. \\

Semantic checking
& Rule, check specification, required evidence,
and other selected rules
& Per-rule judgment, reason, and verbatim
evidence quotations. \\

Appeal review
& Disputed judgment, rule, evidence,
and the agent's counterevidence
& Decision to uphold, overturn, or leave the
judgment uncertain, with justification. \\

Simulated user (learning)
& User profile, task context, interaction
history, and delivered output
& User feedback and internal preference labels
from the four-step decision process. \\

Preference scoring (evaluation)
& Final delivery, public task information,
and preference descriptions
& Per-preference satisfaction, violation,
or uncertainty judgment with evidence. \\

\bottomrule
\end{tabular}
\end{table*}

\begin{table*}[t]
\centering
\small
\setlength{\tabcolsep}{5pt}
\renewcommand{\arraystretch}{1.12}
\caption{Resource limits and runtime settings used in the
experiments in Section~5. Context and prompt sizes are upper bounds.}
\label{tab:tellonce-configuration}
\begin{tabular}{@{}p{0.30\textwidth}p{0.63\textwidth}@{}}
\toprule
Component & Setting \\
\midrule

Auxiliary model
& GPT-6 Luna with low reasoning effort. No sampling parameters are explicitly set. \\

Auxiliary output limit
& 16,384 tokens per call; 8,192 tokens per simulated-user step. \\

Task-solving models
& GPT-5.6 Luna (Codex, medium reasoning);
GPT-5.6 Sol (Codex, low reasoning);
Claude Haiku 4.5 (Claude Code, default settings). \\

Correction-learning context
& Approximately the last 10,000 characters of the session. \\

Conditional context lookup
& At most 5 exact excerpts, totaling up to 8,000 characters. \\

Rule selection
& Up to 32 candidate rules per batch; selected rules
presented in full, up to 24,000 characters. \\

Compiler validation
& 3--9 validation cases per check; up to 3 proposal
revision rounds. \\

Rule checking
& Up to 6 rules per model call; verbatim evidence
quotations required. \\

Prompt limit
& 180,000 characters. \\

Auxiliary request limits
& One retry per request; 900-second timeout. \\

Execution-time checking
& Core-compiled checks through the experimental execution layer. \\

Repair and appeal
& ClawArena and MemoryArena: up to 3 repairs per user input;
SWE-Together: up to 2 repairs per user turn.
One appeal per delivery. \\

\bottomrule
\end{tabular}
\end{table*}

\section{Detailed Experimental Setup}
\label{sec:appendix:experimental-details}

\subsection{Main Experimental Protocol}
\label{sec:appendix:main-protocol}

Our main personalization evaluation covers ClawArena and the travel-planning
tasks from MemoryArena. We evaluate three task-solving models---GPT-5.6 Luna,
GPT-5.6 Sol, and Claude Haiku 4.5---with three fixed user profiles and five
methods: \textsc{No Memory}, \textsc{Mem0}, \textsc{PGMem}, \textsc{Zoro},
and \sysname{}. Each model--user--method--benchmark combination is run
independently. This yields 90 main test groups in total. The four methods with
persistent state are trained separately from an empty long-term store for each
model, user, and benchmark, yielding 72 corresponding learning groups;
\textsc{No Memory} has no cross-task state to train.

\paragraph{Learning phase.}
For each benchmark, methods see the same learning tasks in the same fixed
order. Each persistent-state method starts from an empty store and updates only
its own state. During interaction, the simulated user responds to the behavior
actually produced by that method. Methods therefore share the same feedback
policy and task order, but they are not forced to receive identical correction
text. Each scored unit permits at most one additional simulated-user
correction. The task solver receives the complete natural-language
feedback, while long-term memory updates use only its reusable
preference-bearing content. Feedback that merely restates a
requirement already present in the task is not learned. When
feedback combines task-specific corrections with user preferences,
only the sentences expressing those preferences are used for
long-term updates. The revised delivery is not reviewed again.
Hidden user profiles and simulator rule labels are not exposed.

Training updates only the method's external long-term state. After the final learning
task, the resulting long-term state is frozen before evaluation.

\paragraph{Frozen-state evaluation.}
Test tasks are evaluated without further user teaching. The agent receives only
the benchmark's original inputs; the simulated user does not provide additional
corrections or answer new agent-initiated questions. Persistent-state methods
may read their frozen state throughout evaluation but cannot modify it.

A user-facing delivery may involve multiple model calls, tool operations,
tests, or self-checks before the response is returned. \sysname{} may
additionally select learned rules, evaluate applicable checks, and perform
bounded internal repair. These operations are part of the method and do not
count as new user feedback.

\paragraph{Method comparability.}
All methods retain the solver's normal task context,
tool access, testing facilities, and native self-checking
behavior. Mem0, PGMem, and Zoro retain their original
memory-management mechanisms and prompts, with their
internal model calls using GPT-6 Luna at low reasoning
effort.

During learning, the methods receive the same information
for memory updates: the current task, the agent's preceding
response, and the preference-bearing sentences from user
feedback. At test time, Mem0 retrieves up to ten relevant
memories, limited to 24,000 characters; PGMem retrieves
one passage using its graph-based retriever; and Zoro
provides its knowledge base through its original planning
tools.

Hidden user profiles, reference
annotations, and evaluation results are not provided to the task solver or
memory method, and scores are not returned to the agent during testing.
Task scoring and preference-violation evaluation are
described in Appendix~\ref{sec:appendix:evaluation-scoring}.

\subsection{ClawArena and MemoryArena}
\label{sec:appendix:main-benchmarks}

Our main personalization experiments use ClawArena~\citep{ji2026clawarenabenchmarkingaiagents}
and the \texttt{group\_\allowbreak travel\_\allowbreak planner} configuration of
MemoryArena~\citep{he2026memoryarenabenchmarkingagentmemory}. For both benchmarks, we construct fixed,
researcher-defined splits over complete tasks rather than individual turns.
The splits are deterministic and do not depend on model performance,
preference scores, or the learned memory state.

\paragraph{ClawArena.}
We use 8 complete scenarios for learning and 16 for testing. Starting from the
eligible ClawArena scenarios, we group tasks by the eight original ID-prefix
strata \texttt{c}--\texttt{j}. Within each stratum, scenario IDs are ordered
by a fixed SHA-256 hash key; the first eligible scenario is assigned to
learning and the next two to testing. These prefixes are used only as fixed
selection strata and are not assumed to define semantic task families.

A complete ClawArena scenario consists of the full ordered sequence of rounds
provided for that scenario together with its initial workspace and scheduled
updates. We never place early rounds of one scenario in learning and later
rounds of the same scenario in testing. The 8 learning scenarios contain
240 original questions in total, while the 16 test scenarios contain 474.
Within a scenario, the task session, conversation history, workspace changes,
and previously introduced updates persist across rounds. Independent scenarios
start from fresh execution state; only the method's cross-task long-term state
is carried across scenarios.

We preserve the benchmark's original questions, answer options, answer keys,
round order, and source updates, but execute them through our adapted agent
harness rather than the original OpenClaw runner. The harness makes the public workspace available to
Codex and Claude Code, including the scenario's original
conversation log as a file. Scheduled updates are written
to the workspace before the corresponding round.
For multiple-choice questions, the agent is instructed
to return the selected option letters inside
\verb|\boxed{}|, and the evaluator scores the last such
marker. The harness also implements the benchmark's
execution checks in the isolated evaluation environment. The experimental personalization layer---user
profiles, simulated feedback, long-term memory, and preference scoring---is
added separately from the benchmark's native task criteria.

\paragraph{MemoryArena.}
We restrict MemoryArena to the \texttt{group\_\allowbreak travel\_\allowbreak planner} configuration
and do not interpret the resulting numbers as performance on the full
MemoryArena benchmark. The source contains 270 complete travel-planning rows.
After excluding two rows previously registered for development, we order the
remaining rows using the same fixed hash-based procedure and take the first 90:
the first 30 form the learning set and the next 60 form the test set. This is a
deterministic whole-row split rather than an official MemoryArena split.

Each row is treated as one complete task. It contains an initial traveler
context followed by an ordered sequence of 5--8 travel-planning requests.
The 30 learning tasks contain 211 original questions in total, and the
60 test tasks contain 415. All questions belonging to one row share a
continuous task session and workspace, so earlier requests, delivered plans,
tool context, and task-local feedback remain available within that task.
A new row starts a fresh execution session while the method's cross-task
long-term state remains available according to the protocol in
Appendix~\ref{sec:appendix:main-protocol}.

We preserve the recorded task questions, traveler context, hidden reference
answers, and native travel-scoring criteria, while adapting the environment
to the interactive agent setting. The output template specifies complete travel-plan
submissions and preserves previously established
transportation arrangements, including non-flight
transportation, across requests. The benchmark's public
query tools are exposed through an isolated tool service.
After each delivery, the benchmark's native parser
evaluates all plans submitted so far, using the most
recent plan for each traveler. Hidden answers remain unavailable to the solver.
Native travel-task scoring remains separate from our added user-preference
evaluation.

\paragraph{Shared task identities.}
The same benchmark task identities, split assignments, and learning order are
used across all three solvers, all three user profiles, and all four methods
that maintain long-term state. \textsc{No Memory} uses the same test split but
has no learning phase.

\subsection{Simulated Users and Feedback Generation}
\label{sec:simulated-users}

This appendix describes the profile-based simulated users used in the
ClawArena and MemoryArena experiments, including how user preferences are
represented, how corrections are generated during learning, and the
held-out fidelity study used to validate the simulation procedure.
The general task-improvement simulator used for SWE-Together is described
in Appendix~\ref{sec:appendix:swe-together}.

\subsubsection{User Profile Construction}
\label{sec:simulated-users:construction}

We use three fixed user profiles in the main personalization experiments.
Each profile specifies behavioral preferences together with the conditions
under which they apply and any relevant exceptions. These profiles represent
persistent ways in which a user prefers the agent to behave across tasks,
rather than facts or requirements that belong only to one benchmark instance.

The profiles are private to the simulator. Task-solving agents and memory
methods do not receive the profile, its rule labels, or the simulator's
internal reasoning. They observe only the natural-language feedback that the
simulated user actually sends during interaction.

We also distinguish persistent user preferences from information contained in
the task itself. For example, a preference expressed by a traveler or another
entity inside a benchmark task is not automatically treated as a persistent
preference of the simulated user. Likewise, the initial benchmark request is
not itself treated as a correction-learning event.

\subsubsection{Feedback Generation}
\label{sec:simulated-users:feedback}

During learning, the simulator uses the user's preference
profile and interaction history to decide whether the
agent's behavior warrants correction. It follows four
steps: interpreting the interaction, identifying
applicable preferences, deciding whether to issue a
correction, and checking for overlooked violations.

When a correction is warranted, the simulator generates natural-language
feedback directed at the behavior actually produced by the agent.
If no applicable issue is found, the simulator does not produce a correction.

For the ClawArena and MemoryArena learning runs, each scored unit permits at
most one additional simulated-user correction.  Each memory-based method
starts from its own empty long-term state and learns only from the feedback it
receives.

\subsubsection{Simulator Fidelity}
\label{sec:simulated-users:fidelity}

We evaluate the simulated-user procedure using
held-out interaction data from five AI researchers,
separately from the user profiles used in the main
experiments.

The evaluation contains 790 annotated interaction points: 357 cases in which
the researcher issued a correction and 433 matched cases in which no
correction was given. For each researcher, positive examples are real
correction moments and negative examples are matched non-correction moments.
The simulator is evaluated both on whether it correctly decides to issue a
correction and, for positive examples, whether the generated correction
recovers the issue identified in the human annotation.

The metrics in Table~\ref{tab:sim-user-fidelity} capture two aspects of this
behavior. \emph{Decision precision} measures how often a simulated correction
corresponds to an observed human correction; \emph{decision recall} measures
how often observed correction moments are recovered; \emph{F1} is their
harmonic mean; and \emph{specificity} measures whether the simulator remains
silent at matched non-correction moments. \emph{Rule recall} is stricter: on
positive examples, it measures whether the generated feedback recovers the
human-labeled preference underlying the correction.

\begin{table*}[t]
    \centering
    \small
    \caption{Held-out fidelity evaluation of the simulated-user procedure over
    five real researchers (790 judgments: 357 correction cases and 433
    non-correction cases). Decision metrics evaluate whether the simulator
    identifies when a correction should occur; rule recall evaluates whether
    the generated feedback recovers the human-labeled preference on positive
    examples. Each metric is computed separately for each researcher
and then averaged across the five researchers.}
    \label{tab:sim-user-fidelity}
    \begin{tabular}{lrrrrr}
        \toprule
        & Precision & Recall & F1 & Specificity & Rule recall \\
        \midrule
        Simulator score & 0.864 & 0.953 & 0.906 & 0.940 & 0.668 \\
        \bottomrule
    \end{tabular}
\end{table*}

The simulator achieves an F1 of 0.906 for deciding when to correct and a
specificity of 0.940 on non-correction cases. Recovering the exact
human-labeled preference is more difficult, with a rule recall of 0.668.
These results support the use of simulated feedback while
highlighting the difficulty of recovering users' exact preferences.

\subsection{SWE-Together}
\label{sec:appendix:swe-together}

\paragraph{Task source and split.}
We use the 109 tasks in the fixed SWE-Together task collection
corresponding to the upstream source used by our experiments. The tasks span
26 repository or task-family groups and include repository-level coding
problems with multi-turn user interaction.

We construct a fixed 60-task learning split and 49-task test split. The split
is stratified by repository/task family so that each group contributes to the
learning and test sets according to its size. Within each group, tasks are
ordered by the original interaction time when available, with earlier tasks
used preferentially for learning and later tasks for evaluation. Timestamp
information is available for 100 of the 109 tasks; for the nine Hyperswitch
tasks without a verified interaction time, pull-request order is used as the
ordering proxy.

Each task begins from its own clean repository and independent agent session.
Conversation and repository state are preserved throughout the interaction
within that task, but are not carried into the next task. Cross-task
information is available only through the method's long-term state.

\paragraph{Task-improvement feedback.}
SWE-Together does not use the three personalized user profiles from the main
ClawArena and MemoryArena experiments. Instead, we use a general simulated
user whose role is to help improve completion of the current task. The
simulator follows the task's user intent and interaction history and may
provide additional information, clarify requirements, answer relevant
questions, or point out problems in the current solution.

For \sysname{}, the 60 learning tasks are processed sequentially and feedback
from these interactions can update the long-term rule library.  The
\textsc{No Memory} condition has no cross-task state, but otherwise retains
the same task environment, tools, within-task conversation history, and user
feedback available during the current task.

\paragraph{Multi-turn test interaction.}
Unlike the frozen-feedback protocol used for ClawArena and MemoryArena,
SWE-Together preserves its normal within-task interaction during testing.
After the first user-facing delivery, a task may continue for up to 15
additional interaction turns, for at most 16 user-facing deliveries in total.
These turns may include clarifications, new requirements,
feedback, or instructions to continue execution. The interaction can terminate earlier
when the simulator repeatedly has no additional message to provide.

The long-term state remains frozen throughout these test interactions.
Feedback received within a test task can help the agent continue that task but
does not become cross-task memory for later test tasks. Before each
user-facing delivery, \sysname{} may additionally perform up to two internal
repair steps based on its learned rules.

\paragraph{First and final delivery.}
We evaluate both the first and final user-facing deliveries. The first-delivery score provides a clearer measure of
cross-task experience reuse, as it is obtained before the
agent receives additional feedback on the current test task. The final-delivery score instead measures the quality
reached after the full within-task interaction, during which both
\sysname{} and \textsc{No Memory} can benefit from newly supplied task
information and feedback. All reported SWE-Together results are averaged over three independent runs
using the same task split and evaluation protocol.

Detailed scoring and model configurations are given in
Appendix~\ref{sec:appendix:evaluation-scoring}.

\subsection{Evaluation and Scoring}
\label{sec:appendix:evaluation-scoring}

We evaluate task completion and adherence to learned user requirements
separately. Task performance follows each benchmark's native evaluation,
while preference violations are scored independently from the agent's
user-facing deliveries and produced artifacts. Evaluation results are never
returned to the task-solving agent.

\paragraph{Aggregation across users.}
For ClawArena and MemoryArena, each solver--method combination is evaluated
with three simulated users, each with a distinct preference profile.
We compute each metric separately for each user and report the mean and
sample standard deviation across the three users.

\paragraph{ClawArena task score.}
Each original ClawArena round receives a binary native correctness score.
Multiple-choice rounds are correct only when the predicted answer set matches
the reference answer set. Execution-check rounds are correct when the
task-provided evaluation check succeeds on the delivered workspace.

The 16 test scenarios contain 474 original rounds in total. We pool these
rounds directly and report the percentage that pass:
\[
T_{\mathrm{Claw},u}
=
100\cdot
\frac{\text{number of passed rounds}}{474}.
\]
Every original round receives equal weight, rather than first averaging
scores within scenarios.

\paragraph{MemoryArena task score.}
Each original MemoryArena travel-planning question is evaluated using the
benchmark's native correctness criterion for the plan delivered for that
question. The 60 test tasks contain 415 original questions in total.
We pool their binary native outcomes and report
\[
T_{\mathrm{Memory},u}
=
100\cdot
\frac{\text{number of passed questions}}{415}.
\]
Preference evaluation is performed separately and does not alter the
native task score.

\paragraph{Preference violation rate.}
Preference following is evaluated at the level of individual benchmark
questions, independently of task correctness. For each ClawArena round
or MemoryArena question, the evaluator compares the delivered response
and artifacts against the user's applicable preferences. The evaluator
does not observe the method identity, stored memory, or internal checking
and repair results.

A question is counted as violated if at least one applicable preference
has a confirmed violation, regardless of how many preferences it violates.
For each user $u$ and benchmark $b$, we report
\[
V_{b,u}=\frac{100}{N_b}\sum_{i=1}^{N_b}
\mathbf{1}[\text{unit $i$ is judged violated}],
\]
where $N_{\mathrm{Claw}}=474$ and $N_{\mathrm{Memory}}=415$.

All original questions remain in the denominator. Questions judged
\emph{unknown} or \emph{not applicable} are not counted as violations.

\paragraph{SWE-Together.}
SWE-Together evaluates the first and final user-facing deliveries separately.
Both are scored using the same task-quality criteria and executable repository
evidence. First-delivery scoring uses only information and execution evidence
available by the time of the first delivery; evidence obtained later in the
interaction is not used retroactively. Final-delivery scoring uses the state
available at the end of the complete interaction.

\textsc{No Memory} and \sysname{} are evaluated on the same 49 paired test
tasks for each solver. We perform three independent runs using the same task
split and protocol. Reported bars show the mean across these runs, and
error bars show one sample standard deviation.

%
%

\section{Component Diagnostics}
\label{sec:appendix:component-diagnostics}

The end-to-end experiments evaluate whether \sysname{} ultimately reuses
past corrections effectively. We additionally evaluate several intermediate
decisions in isolation: whether a correction should become persistent,
whether its behavior and scope are extracted correctly, whether later feedback
is reconciled with the appropriate existing rule, whether reliable checks can
be constructed and applied, and whether relevant rules can still be selected
as the library grows.
These diagnostics use separate controlled datasets and denominators.

\subsection{Correction Admission and Rule Extraction}
\label{sec:appendix:component-diagnostics:admission}

\paragraph{Controlled instruction set.}
We construct a held-out set of 192 instructions from 24 recurring behaviors
observed in historical correction logs. Each behavior is expressed under
eight controlled framing variants that vary whether the instruction should
persist, its scope, and whether its activating condition is sufficiently
specified. The resulting set contains 120 instructions that should become
persistent, 48 that should remain task-local, and 24 that express reusable
behavior but leave the activation event underspecified. Gold persistence,
scope, and activation labels are fixed by the framing design before model
evaluation.

The admission prompt is developed separately on 64 examples. On the held-out
set, it correctly rejects all 48 task-local instructions and defers all
24 instructions with underspecified activation, producing no erroneous
persistence decision among these 72 cases. Of the 120 instructions that
should persist, 113 are admitted, corresponding to 94.2\% recall.

Beyond the admission decision, we evaluate whether the model correctly
extracts the rule's scope and activation condition. The complete structural
decision is correct in 175 of 192 cases. For the 120 persistent cases,
two independent semantic judges agree that the extracted behavior, scope,
and activation faithfully preserve the intended requirement in 105 cases.

\begin{table}[htbp]
    \centering
    \small
    \caption{Correction-admission and rule-extraction diagnostics.}
    \label{tab:component-admission}
    \begin{tabularx}{\linewidth}{@{}>{\raggedright\arraybackslash}X >{\raggedleft\arraybackslash}p{0.36\linewidth}@{}}
        \toprule
        Measure & Result \\
        \midrule
        Task-local instructions correctly rejected & 48/48 \\
        Underspecified activation correctly deferred & 24/24 \\
        Erroneous persistence among these cases & 0/72 \\
        Persistent instructions admitted & 113/120 (94.2\%) \\
        Correct decision, scope, and activation & 175/192 (91.1\%) \\
        Semantic fidelity on persistent instructions & 105/120 (87.5\%) \\
        \bottomrule
    \end{tabularx}
\end{table}

\paragraph{Model sensitivity.}
We repeat the same 192-case structural evaluation with three models using
the same prompts and scoring procedure. Gemma 4 31B obtains 175/192 correct
cases, GPT-5.6 Luna obtains 173/192, and Claude Haiku 4.5 obtains 160/192.
Gemma and Luna defer all 24 underspecified-activation cases, while Haiku
does so for 18/24. These results show variation in admission and
extraction accuracy across models.

\begin{table}[htbp]
    \centering
    \small
    \caption{Model sensitivity of the admission and extraction procedure.}
    \label{tab:component-admission-models}
    \begin{tabularx}{\linewidth}{@{}>{\raggedright\arraybackslash}X >{\centering\arraybackslash}X >{\centering\arraybackslash}X@{}}
        \toprule
        Model & Structural accuracy & Underspecified activation deferred \\
        \midrule
        Gemma 4 31B & 175/192 & 24/24 \\
        GPT-5.6 Luna & 173/192 & 24/24 \\
        Claude Haiku 4.5 & 160/192 & 18/24 \\
        \bottomrule
    \end{tabularx}
\end{table}

\subsection{Rule Maintenance and Resolution}
\label{sec:appendix:component-diagnostics:lifecycle}

We evaluate rule maintenance on 16 held-out cases drawn
from real correction logs across three users, covering
restatements, applicability refinements, replacements,
bundled corrections, and new requirements. Each case presents the new
correction together with the existing rules against which it must be resolved,
and the intended update is labeled before evaluation.

The resolver chooses the correct library-update decision in 15/16 cases.
Requiring both the correct update decision and the correct existing rule to be
identified yields 13/16. The remaining errors primarily arise when several
existing rules are semantically related, making the target of the correction
ambiguous.

We also stress-test this resolution process as the
surrounding library grows. Using the same held-out
corrections with libraries containing 2, 47, 197,
and 497 rules, the resolver continues to retrieve
the correct affected rule in every case in which
one exists; update-decision accuracy remains between
13/16 and 15/16 across these library sizes.

\subsection{Check Compilation and Judgment}
\label{sec:appendix:component-diagnostics:checking}

We evaluate this decision using
31 correctly specified rules. Nineteen rules cannot be reliably judged from
the observable runtime event; the compiler correctly declines to construct
an executable check for all 19. Of the remaining 12 checkable rules, 11
produce a check specification and 8 pass all generated positive and negative
validation cases.

Across 176 validation probes for the generated checks, we observe 86 true
positives, 87 true negatives, one false positive, and two false negatives,
for 98.3\% overall agreement with the expected labels.

We separately test the semantic runtime checker on six rules and 48 artifacts.
It produces 23 true positives, 22 true negatives, two false positives, and
one false negative, corresponding to 93.8\% balanced accuracy.

\begin{table}[htbp]
    \centering
    \small
    \caption{Check-compilation and semantic-judgment
results on separate evaluation sets.}
    \label{tab:component-checking}
    \begin{tabularx}{\linewidth}{@{}>{\raggedright\arraybackslash}X >{\raggedleft\arraybackslash}p{0.36\linewidth}@{}}
        \toprule
        Measure & Result \\
        \midrule
        \multicolumn{2}{l}{\textit{Check compilation}} \\
        Uncheckable rules correctly rejected & 19/19 \\
        Checkable rules producing a check & 11/12 \\
        Checks passing all validation cases & 8/12 \\
        Validation probes & 86 TP / 87 TN / 1 FP / 2 FN \\
        Probe-label agreement & 98.3\% \\
        \midrule
        \multicolumn{2}{l}{\textit{Semantic judgment}} \\
        Artifact judgments & 23 TP / 22 TN / 2 FP / 1 FN \\
        Balanced accuracy & 93.8\% \\
        \bottomrule
    \end{tabularx}
\end{table}

\subsection{Rule Selection under Library Growth}
\label{sec:appendix:component-diagnostics:selection}

We evaluate rule selection under library growth using
16 held-out tasks and three shuffled rule orders
at each library size.

\begin{table}[htbp]
    \centering
    \small
    \caption{Task-time rule selection as the rule library grows.}
    \label{tab:component-selection-scaling}
    \begin{tabular}{rcc}
        \toprule
        Rules & Precision & Recall \\
        \midrule
        47  & 84.8\% & 93.3\% \\
        97  & 75.2\% & 86.7\% \\
        197 & 74.6\% & 84.4\% \\
        497 & 70.2\% & 80.0\% \\
        \bottomrule
    \end{tabular}
\end{table}
As the library grows from 47 to 497 rules, recall decreases
from 93.3\% to 80.0\%, while precision decreases from
84.8\% to 70.2\%.

\end{document}